\documentclass{article} % For LaTeX2e
\usepackage{iclr2023_conference,times}

\usepackage{amsmath,amsfonts,bm}

\def\eqref#1{equation~\ref{#1}}
\def\1{\bm{1}}

\def\eps{{\epsilon}}

\def\mL{{\bm{L}}}

\DeclareMathAlphabet{\mathsfit}{\encodingdefault}{\sfdefault}{m}{sl}
\SetMathAlphabet{\mathsfit}{bold}{\encodingdefault}{\sfdefault}{bx}{n}

\newcommand{\E}{\mathbb{E}}

\newcommand{\R}{\mathbb{R}}

\usepackage{cancel}

\usepackage{hyperref}
\usepackage{url}

\usepackage[title]{appendix}

\usepackage[normalem]{ulem} % for crossing out

\usepackage{times}
\usepackage{adjustbox}
\usepackage{amsmath}
\usepackage{changepage}
\usepackage{amssymb}
\usepackage{amsthm}

\usepackage{bbm}
\usepackage{graphicx}%
\usepackage{wrapfig}
\usepackage{makecell}
\usepackage{float}
\usepackage{subfigure}
\usepackage{authblk}
\makeatletter
\newcommand{\printfnsymbol}[1]{%
  \textsuperscript{\@fnsymbol{#1}}}
\makeatother

\usepackage[linesnumbered,ruled,vlined]{algorithm2e}

\newtheorem{theorem}{Theorem}

\newtheorem{innercustomthm}{Theorem}
\newenvironment{customthm}[1]
  {\renewcommand\theinnercustomthm{#1}\innercustomthm}
  {\endinnercustomthm}

\newtheorem{definition}[theorem]{Definition}

\usepackage{color}
\usepackage[utf8]{inputenc} % allow utf-8 input
\usepackage[T1]{fontenc}    % use 8-bit T1 fonts
\usepackage{hyperref}       % hyperlinks
\usepackage{url}            % simple URL typesetting
\usepackage{booktabs}       % professional-quality tables
\usepackage{amsfonts}       % blackboard math symbols
\usepackage{nicefrac}       % compact symbols for 1/2, etc.
\usepackage{microtype}      % microtypography
\usepackage{xspace}
\usepackage{xcolor,colortbl}

\newcommand{\AlgName}{\textsc{Lava}}

\newcommand{\X}{\mathcal{X}}
\newcommand{\Y}{\mathcal{Y}}
\newcommand{\Z}{\mathcal{Z}}
\newcommand{\edit}[1]{\textcolor{black}{#1}}

\newcommand{\add}[1]{\textcolor{black}{#1}}
\newcommand{\nadd}[1]{\textcolor{black}{#1}}

\DeclareMathOperator*{\arginf}{\arg\,\inf}

\newcommand{\ot}{\operatorname{OT}}
\newcommand{\model}{f}
\newcommand{\ft}{f_t}
\newcommand{\fv}{f_v}

\newcommand{\mut}{\mu_t}
\newcommand{\muv}{\mu_v}

\newcommand{\mutft}{\mut^{\ft}}
\newcommand{\muvfv}{\muv^{\fv}}
\renewcommand{\mL}{\mathcal{L}}

\newcommand{\muft}{\mu_{\ft}}
\newcommand{\mufv}{\mu_{\fv}}

\newcommand{\xv}{x_v}
\newcommand{\xt}{x_t}
\newcommand{\yv}{y_v}
\newcommand{\yt}{y_t}

\newcommand{\C}{\mathcal{C}}

\newcommand{\pistar}{\pi^*}
\newcommand{\pistartil}{\widetilde{\pi}^*}
\newcommand{\piy}{\pi_y^*}
\newcommand{\pixy}{\pi_{x, y}^*}
\newcommand{\pixytil}{\widetilde{\pi}_{x, y}^*}

\newcommand{\metric}{\mathrm{d}}

\newcommand{\zeroone}{[0, 1]}
\newcommand{\zeroandone}{\{0, 1\}}

\newcommand{\zeroandoneV}{\{0, 1\}^V}

\title{LAVA: Data Valuation without Pre-Specified Learning Algorithms}
\author[1]{Hoang Anh Just \thanks{
Equal contribution. Repository publicly available on Github: \href{https://github.com/ruoxi-jia-group/LAVA}{https://github.com/ruoxi-jia-group/LAVA}.
}
}
\author[1]{Feiyang Kang
\printfnsymbol{1}}
\author[2]{Jiachen T. Wang}
\author[1]{Yi Zeng}
\author[1]{Myeongseob Ko}
\author[1]{\\Ming Jin}
\author[1]{Ruoxi Jia}
\affil{Virginia Tech, $^2$Princeton University \protect\\
\texttt{\small \{just, fyk, yizeng, myeongseob, jinming, ruoxijia\}@vt.edu  tianhaowang@princeton.edu}}

\iclrfinalcopy % Uncomment for camera-ready version, but NOT for submission.
\begin{document}

\maketitle
\vspace{-2.5em}
\begin{abstract}
\vspace{-1em}
Traditionally, data valuation is posed as a problem of equitably splitting the validation performance of a learning algorithm among the training data. As a result, the calculated data values depend on many \edit{design choices} of the underlying learning algorithm. However, this dependence is undesirable for many use cases of data valuation, such as setting priorities over different data sources in a data acquisition process and informing pricing mechanisms in a data marketplace. In these scenarios, data needs to be valued before the actual analysis and the choice of the learning algorithm is still undetermined then. Another side-effect of the dependence is that to assess the value of individual points, one needs to re-run the learning algorithm with and without a point, which incurs a large computation burden. 
\vspace{0.1em}

This work leapfrogs over the current limits of data valuation methods by introducing a new framework that can value training data in a way that is oblivious to the downstream learning algorithm. Our main results are as follows. \textbf{(1)} We develop a proxy for the validation performance associated with a training set based on a \edit{non-conventional \textit{class-wise}} \textit{Wasserstein distance} between the training and the validation set. We show that the distance characterizes the upper bound of the validation performance for any given model under certain Lipschitz conditions. \textbf{(2)} We develop a novel method to value individual data based on the sensitivity analysis of \edit{the class-wise} Wasserstein distance. \edit{Importantly, these values can be directly obtained \emph{for free} from the output of off-the-shelf optimization solvers when computing the distance.} \textbf{(3) }We evaluate our new data valuation framework over various use cases related to detecting low-quality data
and show that, surprisingly, the learning-agnostic feature of our framework enables a significant improvement over the state-of-the-art performance while being orders of magnitude faster. 
\end{abstract}
\vspace{-2em}

\section{Introduction} \label{intro}
% \vspace{-0.5em}
\vspace{-1em}
Advances in machine learning (ML) crucially rely on the availability of large, relevant, and high-quality datasets. However, real-world data sources often come in different sizes, relevance levels, and qualities, differing in their value for an ML task. Hence, a fundamental question is how to quantify the value of individual data sources. Data valuation has a wide range of use cases both within the domain of ML and beyond. It can help practitioners enhance the model performance through prioritizing high-value data sources \citep{ghorbani2019data}, and 
%is also central to the emerging data economy as 
it allows one to make strategic and economic decisions in data exchange~\citep{dataeconomy}. 

% as to which data would be the most beneficial to acquire. 

In the past literature~\citep{ghorbani2019data,jia2019towards,kwon2021beta}, data valuation is posed as a problem of equitably splitting the validation performance of a given learning algorithm among the training data. Formally, given a training dataset $\mathbb{D}_t=\{z_i\}_{i=1}^N$, a validation dataset $\mathbb{D}_v$, a learning algorithm $\mathcal{A}$, and a model performance metric $\textsc{perf}$ (e.g., classification accuracy), a \emph{utility function} is first defined over all subsets $S\subseteq \mathbb{D}_\text{t}$ of the training data: $U(S) := \textsc{perf}(\mathcal{A}(S))$. \edit{Then,} the objective of data valuation is to find a score vector $s\in \mathbb{R}^N$ that represents the allocation to each datapoint. For instance, one simple way to value a point \edit{$z_i$} is through leave-one-out (LOO) error $U(\mathbb{D}_t)-U(\mathbb{D}_t\setminus \{z_i\})$, i.e., the change of model performance when the point is excluded from training. Most of the recent works have leveraged concepts originating from cooperative game theory (CGT), such as the Shapley value~\citep{ghorbani2019data,jia2019towards}, Banzhaf value \citep{wang2022data}, general semivalues~\citep{kwon2021beta}, and Least cores~\citep{yan2021if} to value data. Like the LOO, all of these concepts are defined based on the utility function. 
% Specifically, the Shapley value and semi-values score a point based on some weighted average of the utility change caused by the addition of the point. Least cores and Nucleolus, on the other hand, optimizes the scores under the constraints of discrepancy of cumulative scores $\sum_{i\in S} s_i$ over any set $S$ and its corresponding utility $U(S)$.

Since the utility function is defined w.r.t. a specific learning algorithm, the data values calculated from the utility function also depend on the learning algorithm. In practice, there are many choice points pertaining to a learning algorithm, such as the model to be trained, the type of learning algorithm, as well as the hyperparameters. The detailed settings of the learning algorithms are often derived from data analysis. However, in many critical applications of data valuation such as informing data acquisition priorities and designing data pricing mechanism, data needs to be valued before the actual analysis and the choice points of the learning algorithm are still undetermined at that time. This gap presents a main hurdle for deploying existing data valuation schemes in the real world.

The reliance on learning algorithms also makes existing data valuation schemes difficult to scale to large datasets. The exact evaluation of LOO error and CGT-based data value notions require evaluating utility functions over different subsets and each evaluation entails retraining the model on that subset: the number of retraining times is linear in the number of data points for the former, and exponential for the latter. While existing works have proposed a variety of approximation algorithms, scaling up the calculation of these notions to large datasets remains expensive. 
%\edit{Further, model-based approaches crucially rely on the performance scores associated with models trained on different subsets to determine the value of data, naturally susceptible to noise and the accuracy of the validation results is often affected~\citep{wang2022data}, while “noise-tolerant” modeling (e.g., overparameterized DNNs~\citep{rolnick2017deep}) may cause the model performance to be less sensitive to problematic datapoints, posing additional challenges for building data valuation applications (see ~\ref{ap-data_based}).}
% \tianhao{[the second half of the sentence is wrong]}
\edit{Further, learning-algorithm-dependent approaches rely on the performance scores associated with models trained on different subsets to determine the value of data; thus, they are susceptible to noise due to training stochasticity when the learning algorithm is randomized (e.g., SGD) \citep{wang2022data}.
% Additionally, different models have varying degrees of robustness against bad data~\citep{rolnick2017deep}, posing additional challenges for building data valuation applications.
}

This work addresses these limitations by introducing a \emph{\textbf{l}earning-\textbf{a}gnostic} data \textbf{va}luation (\AlgName) framework. $\AlgName$ is able to produce efficient and useful estimates of data value in a way that is oblivious to downstream learning algorithms. Our technical contributions are listed as follows.

\textbf{Proxy for validation performance.} We propose a proxy for the validation performance associated with a training set based on the \edit{non-conventional class-wise} Wasserstein distance~\citep{alvarez2020geometric} between the training and the validation set. The \edit{hierarchically-defined} Wasserstein distance utilizes a hybrid Euclidean-Wasserstein cost function to compare the feature-label pairs across datasets. We show that this distance characterizes the upper bound of the validation performance of any given models under certain Lipschitz conditions. 
%   This result provides theoretical justification for using this distance as a proxy for validation performance in data valuation.

\textbf{Sensitivity-analysis-based data valuation.} We develop a method to assess the value of an individual training point by analyzing the sensitivity of the \edit{particular} Wasserstein distance to the perturbations on the corresponding probability mass. \edit{
The values can be directly obtained \emph{for free} from the output of off-the-shelf optimization solvers once the Wasserstein distance is computed.} As the Wasserstein distance can be solved much more efficiently with entropy regularization~\citep{cuturi2013sinkhorn}, in our experiments, we utilize the duals of the entropy-regularized program to approximate the sensitivity. Remarkably, we show that the gap between two data values under the \edit{original} non-regularized Wasserstein distance can be recovered \emph{exactly} from the solutions to the regularized program.
% This bound can be used to certify the consistency between the groundtruth data value rankings and the approximate ones.

\textbf{State-of-the-art performance for differentiating data quality.} We evaluate $\AlgName$ over a wide range of use cases, including detecting mislabeled data, backdoor attacks, poisoning attacks, noisy features, and task-irrelevant data, in which some of these are first conducted in the data valuation setting. Our results show that, surprisingly, the learning-agnostic feature of our framework enables a significant performance improvement over existing methods, while being orders of magnitude faster.

\vspace{-0.5em}
\section{Measuring Dataset Utility via Optimal Transport} \label{dataset_val}
\vspace{-0.5em}
In this section, we consider the problem of quantifying training data utility $U(\mathbb{D}_t)$ without the knowledge of learning algorithms. Similar to most of the existing data valuation frameworks, we assume access to a set of validation points $\mathbb{D}_v$. Our idea is inspired by recent work on using the \edit{\emph{hierarchically-defined Wasserstein distance}} to characterize the relatedness of two datasets~\citep{alvarez2020geometric}. Our contribution here is to apply that \edit{particular} Wasserstein distance to the data valuation problem and provide a theoretical result that connects the distance to validation performance of a model, which might be of independent interest.

\vspace{-0.5em}
\subsection{Optimal Transport-based Dataset Distance}
\vspace{-0.5em}
\textbf{Background on Optimal Transport (OT).} OT is a celebrated choice for measuring the discrepancy between probability distributions~\citep{villani2009optimal}.
% racing back to a line of well-established research in statistics (\citet{monge1781memoire}, \citet{kantorovich1942translocation}), the well-studied approach emerges as a powerful tool for its desirable analytical properties (\citet{genevay2018learning}). 
Compared to other notable dissimilarity measures such as the Kullback-Leibler Divergence~\citep{kullback1951information} or Maximum Mean Discrepancies (MMD)~\citep{szekely2005hierarchical}\add{, the mathematically well-defined OT distance has advantageous analytical properties. For instance, OT is a distance metric, 
%(is a valid metric; compatible with sparse-support distributions; stable with respect to deformations of the distributions’ supports), 
being computationally tractable and computable from finite samples \citep{ genevay2018learning, feydy2019interpolating}.}
% (smoothness, probability space, etc.) (\citet{genevay2018learning}). Gaining popularity in modern ML, progresses from the active research area have been consistently extending its fields of applications, such as in domain adaptation/transfer learning (\citet{flamary2016optimal}), generative model learning (\citet{goodfellow2014generative}), computer graphics (\citet{bonneel2016wasserstein}), etc.

The Kantorovich formulation~\citep{kantorovich1942translocation} defines the OT problem as a Linear Program (LP). 
Given probability measures $\mu_t, \mu_v$ over the space $\Z$, 
the OT problem is defined as
$
\label{OT}
\ot(\mu_t, \mu_v) := \min_{\pi \in \Pi(\mu_t, \mu_v)} \int_{\mathcal{Z}^2} \add{\C}(z, z') d \pi(z, z')
$ \label{eq:ot}
where $\label{primal-const}
    \Pi(\mu_t, \mu_v) :=\left\{\pi \in \mathcal{P}(\add{\Z \times \Z})\mid \int_\Z \pi\add{(z, z')} dz=\mu_t, \int_{\add{\Z}} \pi\add{(z, z')} dz'=\mu_v\right\}
$ denotes a collection of couplings \add{between two distributions} $\mu_t$ and $\mu_v$ 
\add{and} $\add{\C: \Z \times \Z} \rightarrow \mathbb{R}^{+}$ is some symmetric positive cost function (with $\add{\C}(z, z)=0$), respectively. 
\edit{If $\C(z, z')$ is the Euclidean distance between $z$ and $z'$ (i.e., $\ell^2$-norm) according to the distance metric $\metric$, then $\ot(\mu_t, \mu_v)$ is 2-Wasserstein distance, which we denote as $W_\C(\mu_t, \mu_v) = W_\metric(\mu_t, \mu_v) := \ot(\mu_t, \mu_v)$.}
{In this work, the notation OT and $W$ are used interchangeably, with a slight difference that we use OT to emphasize various of its formulations while $W$ \edit{specifies on which distance metric it is computed.}}

% When the cost function $c(z, z')$ is the $p$-norm on the metric space $\mathcal{Z}\subset \mathbb{R}^D$ for some $p \geq 1$,
% $
% c(x,y)=\|x-y\|^p,
% $

%Given a metric $d_\mathcal{Z}$ equipped with $\Z$, it is natural to set $c(z,z')=d_\mathcal{Z}(z,z')^p$ for some $p \geq 1$. The resulting OT distance $\operatorname{OT}(\mu_t, \mu_v)^{1/p}$ is referred to as the $p$-Wasserstein distance, denoted by $W_p(\mu_t, \mu_v)$.

% The case $p=1$ is also known as the Earth Mover's distance (\citet{rubner2000earth}), and $p=2$ gives the commonly used quadratic Wasserstein.

% The marginal distributions $\mu_t, \mu_v$ is rarely known exactly in practice. Instead, the  is often defined on discrete distributions that are empirically approximations to the original contious distributions based on finite samples (\citet{alvarez2020geometric}). The closed-form of the Wasserstein distance offers one the major advantages in its applications.

\textbf{Measuring Dataset Distance.} 
We consider a multi-label setting where we denote \edit{$\ft: \X \rightarrow \zeroandoneV$, $\fv: \X \rightarrow \zeroandoneV$} as the labeling functions for training and validation data, respectively, \edit{where $V$ is the number of different labels.}
Given the training set $\mathbb{D}_t=\add{\{(x_i, \ft(x_i))\}_{i=1}^{N}}$ \add{of size $N$}, and the validation set $\mathbb{D}_v=\add{\{(x_i', \fv(x_i'))\}_{i=1}^{M}}$ \add{of size $M$}, one can construct discrete measures $\mu_t(x,y) := \add{\frac{1}{N}} \sum_{i=1}^N \delta_{\add{(x_i,y_i)}}$ and $\mu_v(x,y) := \add{\frac{1}{M}} \sum_{i=1}^M \delta_{\add{(x_i',y_i')}}$\add{, where $\delta$ is Dirac function}. 
%The distance between the two datasets can be calculated by the OT distance between the two discrete measures. 
% \subsection{Distance between datasets}
% In the context of domain adaptation and transfer learning, the utility of a given dataset on a target task is often measured by the distribution discrepancy between the given dataset and the target dataset.
% Datasets for supervised learning are typically consisting of feature-label pairs.
Consider that each datapoint consists of a feature-label pair $(x_i,y_i) \in \X \times \Y$. While the Euclidean distance naturally provides the metric to measure distances between features, the distance between labels generally lacks a definition. 
% The sets of labels for different datasets might be inconsistent, raising prominent challenges for the application of OT. Notable progress on this issue has been made in \citep{alvarez2020geometric}. The authors~
% Inspired by~\citep{alvarez2020geometric}, we measure the distance between two labels in terms of the OT distance between the conditional distribution of the features given each label. 
% \citep{alvarez2020geometric} demonstrate the usefulness of this definition in transfer learning applications. 
Consequently, we define conditional distributions $\mut(x|y) := \frac{ \mut(x)I[\ft(x)=y] }{ \int \mut(x)I[\ft(x)=y] dx }$ and $\muv(x|y) := \frac{ \muv(x)I[\fv(x)=y] }{ \int \muv(x) I[\fv(x)=y] dx }$.
Inspired by \cite{alvarez2020geometric}, we measure the distance between two labels in terms of the OT distance between the conditional distributions of the features given each label. Formally, we adopt the following cost function between feature-label pairs:
$
\label{eqn:hybrid_cost}
    \C( (x_{t}, y_{t}), (x_{v}, y_{v}) ) := \metric(x_{t}, x_{v}) + c W_{\metric}(\mut(\cdot | y_t),  \muv(\cdot | y_v)   ),
$
\add{where $c \ge 0$ is a weight coefficient.
We note that $\C$ is a distance metric since $W_\metric$ is a valid distance metric.} With the definition of $\C$, we propose to measure the distance between the training and validation sets using the \edit{non-conventional, hierarchically-defined} Wasserstein distance between the corresponding discrete measures:
$
W_\C\left(\mu_t,\mu_v\right)=\min _{\pi \in \Pi(\mu_t, \mu_v)} \int_{\edit{\mathcal{Z}^2}} \C \left(z, z^{\prime}\right) d \pi\left(z, z^{\prime}\right).$

\edit{Despite its usefulness and potentially broad applications, we note that it remains absent for existing research to explore its theoretical properties or establish applications upon this notion. This work aims to fill this gap by extending in both directions–novel analytical results are presented to provide its theoretical justifications while an original computing framework is proposed that extends its applications to a new scenario of datapoint valuation.}

\textbf{Computational Acceleration via Entropic Regularization.} Solving the problem above scales cubically with $MN$, which is prohibitive for large datasets. Entropy-regularized OT (entropy-OT) becomes a prevailing choice for approximating OT distances as it allows for fastest-known algorithms. Using the iterative Sinkhorn algorithm~\citep{cuturi2013sinkhorn} with almost linear time complexity and memory overhead, entropy-OT can be implemented on a large scale with parallel computing~\citep{genevay2018learning,feydy2019interpolating}.
% \subsection{Entropy regularized }
%  enjoys many desirable analytical properties, though, it has been plagued by computational issues. Formulated as a Linear Program, the state of art solution algorithms have a time complexity of $\mathcal{O}(n^3 \log(n))$ while also suffering from the curse of dimensionality (\citet{genevay2018learning}), making it impossible for the scale of ML.
% Attracting growing attention in recent years, entropy regularization emerges as an effective approach for linear programming (\citet{weed2018explicit}). Providing crucial foundations for establishing efficient algorithms with tractable analytical properties, entropy-regularized  \textit{(entropy-OT)} becomes a prevailing choice for approximating OT distances as it allows for fastest-known algorithms. Leveraging the iterative Sinkhorn algorithm (\citet{cuturi2013sinkhorn}) with almost linear time complexity and memory overhead, entropy-OT can be implemented on a large scale with parallel computing and has provable convergence (\citet{genevay2018learning}, \citet{feydy2019interpolating}).
Given a regularization parameter, $\varepsilon >0$, entropy-OT can be formulated as 
$\label{EOT}
    \operatorname{OT}_\varepsilon(\mu_t, \mu_v) \,\,:= \min_{\pi \in \Pi(\mu_t, \mu_v)} \int_{\edit{\Z^2}} \C(\edit{z,z'}) d \pi(\edit{z,z'})+\varepsilon H(\pi|\mu_t\otimes\mu_v),
$ \label{eq:ot_eps}
where $H(\pi|\mu_t\otimes\mu_v)= \int_{\Z^2} \log\left( \frac{d\pi}{d\mu_t d\mu_v}\right) d\pi$.
% the penalization term $H(\pi|\mu_t\otimes\mu_v)$ can be defined as the \textit{Kullback-Leibler (KL)} divergence 
% \begin{equation*}
% H(\pi|\mu_t\otimes\mu_v)=\operatorname{KL}(\pi|\mu_t\otimes\mu_v):= \int_{\X^2} \log\left( \frac{d\pi}{d\mu_t d\mu_v}\right) d\pi
% \end{equation*}
As $\varepsilon\rightarrow0$, the dual solutions to the $\varepsilon$-entropy-OT converge to its OT counterparts as long as the latter are unique \citep{nutz2021entropic}. 
%in $L^1$ as soon as the latter are unique \citep{nutz2021entropic}. 
% introduce  distance

% introduce the theorem about the connection of this distance and the model performance
\vspace{-1em}
\subsection{Smaller \edit{\emph{Class-Wise Wasserstein Distance} Entails Better Validation Performance}}
\vspace{-0.5em}
% Having demonstrated the analytical convenience and implementation capabilities of the , we finally establish the theoretical foundation of using the Wasserstein distance as a proxy for the dataset utility. With mild assumptions on Lipschitzness of the downstream model as well as the labeling functions associated with the training and validation set, we show that the discrepancy between the training and validation performance of a model is bounded by the Wasserstein distance between the training and the validation dataset, providing theoretical guarantees for using  distance to predict the dataset utility (i.e., validation performance) without the knowledge of the downstream models.

\begin{wrapfigure}{R}{0.35\columnwidth}
    \centering
    \vspace{-1em}
    \includegraphics[width=0.35\columnwidth]{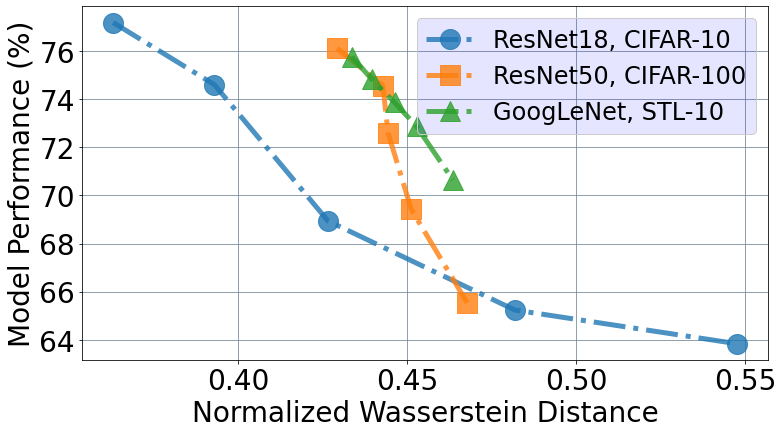}
        \vspace{-1.5em}
    \caption{Normalized Wasserstein distance vs. model performance on different datasets and models.
    }
        \vspace{-1em}
    \label{fig:theory}
\end{wrapfigure}

In this paper, we propose to use $W_\C$, a \emph{\edit{non-conventional, class-wise} Wasserstein distance} w.r.t. the special distance function $\C$ defined in  \ref{eqn:hybrid_cost}%\kang{Emphasis on the unique contribution of this section: DISTINGUISH from similar-looking existing results on Wasserstein distance and domain transfer and underscore the novelty of what is going to be presented here.}
, as a learning-agnostic surrogate of validation performance to measure the utility of training data. 
Note that while Wasserstein distances have been frequently used to bound the learning performance change due to distribution drift \citep{courty2017joint, damodaran2018deepjdot, shen2018wasserstein, ge2021ota}, % \tianhao{better to have citations here, e.g. Shen2018 AAAI, Courty NIPS17, and more recent works maybe; also I think it's good to move this sentence to the beginning of the subsection since people might lose interest soon after they see the title; or, we should change the title to `class-wise wasserstein distance' or something}
 this paper is the first to bound the performance change by the \edit{hierarchically-defined} Wasserstein distance with respect to the hybrid cost $\C$.

Figure~\ref{fig:theory} provides an empirical justification for using this \edit{novel} %\kang{Is this enough to highlight that this is a new result?}
distance \edit{metric} as a proxy, and presents a relation between the \edit{class-wise} Wasserstein distance and a model's validation performance. Each curve represents a certain dataset trained on a specific model to receive its performance. Since, each dataset is of different size and structure, their distances will be of different scale. Therefore, we normalize the distances to the same scale to present the relation between the Wasserstein distance and model performance, which shows that despite different datasets and models, with increased distance, the validation performance decreases.

The next theorem theoretically justifies using \add{this} Wasserstein distance as a proxy for validation performance of a model. 
With assumptions on Lipschitzness of the downstream model as well as the labeling functions associated with the training and validation sets \add{(as explicated in Appendix~\ref{appendix:proof-bound})}, we show that the discrepancy between the training and validation performance of a model is bounded by the \edit{hierarchically-defined} Wasserstein distance between the training and the validation datasets. 

\begin{theorem}
\label{thm:bound}
\add{We denote $\ft: \X \rightarrow \zeroandone$, $\fv: \X \rightarrow \zeroandone$ as the labeling function for training and validation data. Let $\model: \X \rightarrow \zeroone$ be the model trained on training data. 
Let $\mut$, $\muv$ be the training and validation distribution respectively, and let $\mut(\cdot|y)$ and $\muv(\cdot|y)$ be the corresponding conditional distribution given label $y$. 
% \add{We define conditional distribution $\mut(x|y) := \frac{ \mut(x)I[\ft(x)=y] }{ \int \mut(x)I[\ft(x)=y] dx }$ and $\muv(x|y) := \frac{ \muv(x)I[\fv(x)=y] }{ \int \muv(x) I[\fv(x)=y] dx }$. }
Assume that there exists constant $M$ such that $|\model(x)|, |\ft(x)|, |\fv(x)| \le M$ for all $x \in \X$, the model $f$ is $\eps$-Lipschitz, and the loss function $\mL: \zeroone \times \zeroone \rightarrow \R^+$ is $k$-Lipschitz in both inputs. 
Define cost function $\C$ between $(x_v, y_v)$ and $(x_t, y_t)$ as }
\begin{align}
    \add{\C}( (x_t, y_t), (x_v, y_v) ) := \add{\metric}(\xt, \xv) + \add{c} W_{\add{\metric}}( \add{\mut(\cdot | y_t)}, \add{\muv(\cdot | y_v)}  ),
\end{align}
\add{where $c$ is a constant}. \add{Under a certain cross-Lipschitzness assumption for $\ft$ and $\fv$ detailed in Appendix \ref{appendix:proof-bound}, we have }
\begin{align}
    \E_{\add{x \sim \muv(x)}} \left[ \add{\mL}(\add{\fv(x)}, \model(x)) \right]
    \le \E_{\add{x \sim \mut(x)}} \left[ \add{\mL}(\add{\ft(x)}, \model(x)) \right] + \add{k\eps}W_{\add{\C}}( \add{\mu_t}, \add{\mu_v}) + \add{\mathcal{O}(kM)}.
\end{align}
\end{theorem}

Proofs are deferred to Appendix~\ref{appendix:proof-bound}. The bound is interesting to interpret. The first term on the right-hand side corresponds to the training performance. In practice, when a model with large enough capacity is used, this term is small. The second one is the exact expression of the Wasserstein distance that we propose to use as a proxy for validation performance. The last error term is due to possible violation of the cross-Lipschitzness assumption for \edit{$f_t$} and $f_v$. This term will be small if \edit{$f_t$} and $f_v$ assign the same label to close features with high probability. If the last term is small enough, it is possible to use the proposed Wasserstein distance as proxy for validation loss provided that $f$, $f_t$ and $f_v$ verify the cross-Lipschitz assumptions. The bound resonates with the empirical observation in Figure~\ref{fig:theory} that with lower distance between the training and the validation data, the validation loss of the trained model decreases.

% This provides an empirical justification for Theorem~\ref{thm:bound}.

% \textbf{Wasserstein Distance vs Model Performance.} 

% talk about properties of this data utility metric

\vspace{-0.5em}
\section{Efficient Valuation of Individual Datapoints}
\vspace{-0.5em}\label{data_val}
Note that the \edit{class-wise} Wasserstein distance defined in the previous section can be used to measure the utility for subsets of $\mathbb{D}_t$. Given this utility function, one can potentially use existing CGT-based notions such as the Shapley value to measure the contribution of individual points. However, even approximating these notions requires evaluating the utility function on a large number of subsets, which incurs large extra computation costs. In this section, we introduce a new approach to valuating individual points. \edit{Remarkably, our values can be directly obtained \emph{for free} from the output of off-the-shelf optimization solvers once the \edit{proposed} Wasserstein distance between the full training and testing datasets is computed.}
\vspace{-0.5em}
\subsection{Datapoint Valuation via Parameter Sensitivity}
OT distance is known to be insensitive to small differences while also being not robust to large deviations~\citep{villani2021topics}.
% –the OT distance will go to $\infty$ if the some distance between datapoints goes to  $\infty$~\citep{villani2021topics}. 
This feature is naturally suitable for detecting abnormal datapoints---disregarding normal variations in distances between clean data while being sensitive to abnormal distances of outlying points. We propose to measure individual points' contribution based on the gradient of the OT distance to perturbations on the probability mass associated with each point.
% Leveraging the parameter sensitivity of the OT LP problem, we develop an approach to measuring the contribution of each data point to the OT distance and this gradient is proposed to be used as a metric for data valuation. 

% LP naturally lends itself to this kind of sensitivity analysis (\citet{bertsimas1997introduction}). 
Gradients are local information. However, unlike widely used influence functions that only hold for infinitesimal perturbation~\citep{koh2017understanding}, gradients for LP hold precisely in a local range and still encode partial information beyond that range, making it capable of reliably predicting the change to the OT distance due to adding or removing datapoints without the need of re-calculation. Also, the gradients are directed information, revealing both positive and negative contributions for each datapoint and allowing one to perform ranking of datapoints based on the gradient values. Finally, the OT distance always considers the collective effect of all datapoints in the dataset. 
Leveraging the duality theorem for LP, we rewrite the \emph{original} OT problem (introduced in \ref{OT}) in the equivalent form:
$\label{DOT}
%\begin{aligned}
    \operatorname{OT}(\mu_t, \mu_v) :=
    \max_{(f,g)\in C^0(\mathcal{Z})^2}\langle f, \mu_t\rangle + \langle  g, \mu_v\rangle,
%\end{aligned}
$
where $C^0(\mathcal{Z})$ is the set of all continuous functions, 
% $\langle f, \mu_t\rangle\stackrel{\text{def}}{=}\int_\mathcal{Z}f d\mu_t=\mathbb{E}_{\mu_t} (f)$ and $\langle g, \mu_v\rangle\stackrel{\text{def}}{=}\int_\mathcal{Z}g d\mu_t=\mathbb{E}_{\mu_v} (g)$, 
$f$ and $g$ are the dual variables. Let $\pi^*$ and $(f^*, g^*)$ be the corresponding optimal solutions to the primal and dual problems. The Strong Duality Theorem indicates that $\operatorname{OT}(\pi^*(\mu_t,\mu_v))=\operatorname{OT}(f^*,g^*)$, where the right-hand side is the  distance parameterized by $\mu_t$ and $\mu_v$. From the Sensitivity Theorem~\citep{bertsekas1997nonlinear}, we have that the gradient of the  distance w.r.t. the probability mass of datapoints in the two datasets can be expressed as follows:
$
\nabla_{\mu_t} \operatorname{OT}(f^*, g^*)=(f^*)^T,\,\, \nabla_{\mu_v} \operatorname{OT}(f^*, g^*)=(g^*)^T.
$
% Let $\pi^*$ and $(f^*, g^*)$ be the corresponding optimal solutions to the primal and dual problems. The Strong Duality Theorem gives that $\operatorname{OT}(\pi^*(A, B))=\operatorname{OT}(f^*, g^*)$, where the latter is the  distance parameterized by $\text{prob}(A)$ and $\text{prob}(B)$. From the Sensitivity Theorem (\citet{bertsekas1997nonlinear}), we have
% \begin{equation*}
% \nabla_{\text{prob}(A)} \operatorname{OT}(f^*, g^*)=(f^*)',\quad \nabla_{\text{prob}(B)} \operatorname{OT}(f^*, g^*)=(g^*)'
% \end{equation*}
% which is the gradient of the optimal transport distance w.r.t. the probability mass of datapoints in the two datasets.
% Note that the formulation in Eq.~\ref{DSOT} is always redundant as the constraint $\sum_{i=1}^m \text{prob}(a_i) = \sum _{i=1}^n \text{prob}(b_i) = 1$ is already implied. This renders the dual solution to be non-unique. To address this issue, we first remove any one of the constraints in Eq.~\ref{DSOTCS} and make the primal formulation non-degenerate. Then, we assign a value of zero to the dual variable corresponding to the removed primal constraint.
Note that the original formulation in ~\ref{OT} is always redundant as the constraint $\sum_{i=1}^N \mu_t(z_i) = \sum _{i=1}^M \mu_v(z_i') = 1$ is already implied, rendering the dual solution to be non-unique. To address this issue, we first remove any one of the constraints in $\Pi(\mu_t, \mu_v)$ and make the primal formulation non-degenerate. Then, we assign a value of zero to the dual variable corresponding to that removed primal constraint.

% When measuring the gradients of the optimal transport distance w.r.t. to the probability mass of datapoints in each datasets, we calculate the \textit{calibrated gradient} as 
% \begin{equation}\label{grd}
%     \frac{\partial\operatorname{OT(A,B)}}{\partial\operatorname{prob}(a_i)} = f^*_i-\sum_{j\in\{1, ...m\}\setminus i} \frac{f^*_j}{m-1},\quad \frac{\partial\operatorname{OT(A,B)}}{\partial\operatorname{prob}(b_i)} = g^*_i-\sum_{j\in\{1, ...n\}\setminus i} \frac{g^*_j}{n-1}
% \end{equation}
% which represents the rate of change in the optimal transport distance w.r.t the change of the probability mass of a datapoint along the direction ensuring the probability mass for all datapoints in the dataset always sum up to 1. The value of calibrated gradients is independent of the choice of selection during the constraint removal.

When measuring the gradients of the OT distance w.r.t. the probability mass of a given datapoint in each dataset, we calculate the \textit{calibrated gradient} as \vspace{-0em}
\begin{equation}\label{grd}
    \frac{\partial\operatorname{OT}(\mu_t,\mu_v)}{\partial\mu_t(z_i)} = f^*_i-\sum_{j\in\{1, ...N\}\setminus i} \frac{f^*_j}{N-1},\quad \frac{\partial\operatorname{OT}(\mu_t,\mu_v)}{\partial\mu_v(z_i')} = g^*_i-\sum_{j\in\{1, ...M\}\setminus i} \frac{g^*_j}{M-1},
\end{equation} 

\vspace{-1em}
\edit{which represents the rate of change in the OT distance w.r.t the change of the probability mass of a given datapoint \emph{along the direction} ensuring the probability mass for all datapoints in the dataset always sums up to one (explicitly enforcing the removed constraint)}. %\hoang{[this sentence is a bit long and breaking up might be helpful]} 
The value of calibrated gradients is independent of the choice of selection during the constraint removal.

% \begin{remark}[Datapoint valuation via calibrated gradients]
\textbf{Datapoint valuation via calibrated gradients.}
% Derived from the sensitivity theorems in Linear Programming, t
The calibrated gradients predict how the OT distance changes as more probability mass is shifted to a given datapoint. This can be interpreted as a measure of the contribution of the datapoint to the OT distance. The contribution can be positive or negative, suggesting shifting more probability mass to this datapoint would result in an increase or decrease of the dataset distance, respectively. If we want a training set to match the distribution of the validation dataset, then removing datapoints with large positive gradients while increasing datapoints with large negative gradients can be expected to reduce their OT distance. As we will show later, 
the calibrated gradients can provide a tool to detect abnormal or irrelevant data in various applications.
% measure the contribution of each datapoint to the discrepancy between datasets and evaluate the quality of datapoints in various applications.
% \end{remark}
% \begin{remark}[Radius for accurate predictions] 

\textbf{Radius for accurate predictions.}
The Linear Programming theories~\citep{bertsimas1997introduction} give that for each non-degenerate optimal solution, we are always able to perturb parameters on the right-hand side of primal constraints \edit{($\Pi(\mu_t, \mu_v)$ in ~\ref{primal-const})} in a small range without affecting the optimal solution to the dual problem. 
% As the optimal objective value is the sum of optimal dual solutions weighted by these parameters, the optimal dual solution vector can be interpreted as the marginal cost in various applications. 
When the perturbation goes beyond a certain range, the dual solution becomes primal infeasible and the optimization problem needs to be solved again. 
% In the context of OT, the calibrated gradients we calculated are the marginal cost of the OT distance per unit increase of the probability mass of a datapoint (while uniformly decreasing the probability mass of all other datapoints in the dataset such that all probability mass always sums up to 1). 
Hence, the calibrated gradients are local information and we would like to know the perturbation radius such that the optimal dual solution remains unchanged---i.e., whether this range is large enough such that the calibrated gradients can accurately predict the \emph{actual} change to the OT distance. If the perturbation goes beyond this range, the prediction may become inaccurate as the dual solution only encodes partial information about the optimization. 

\begin{figure*}
% \vspace{-2.5em}
    \centering%
	\includegraphics[height=3.3cm,width=0.92\linewidth]{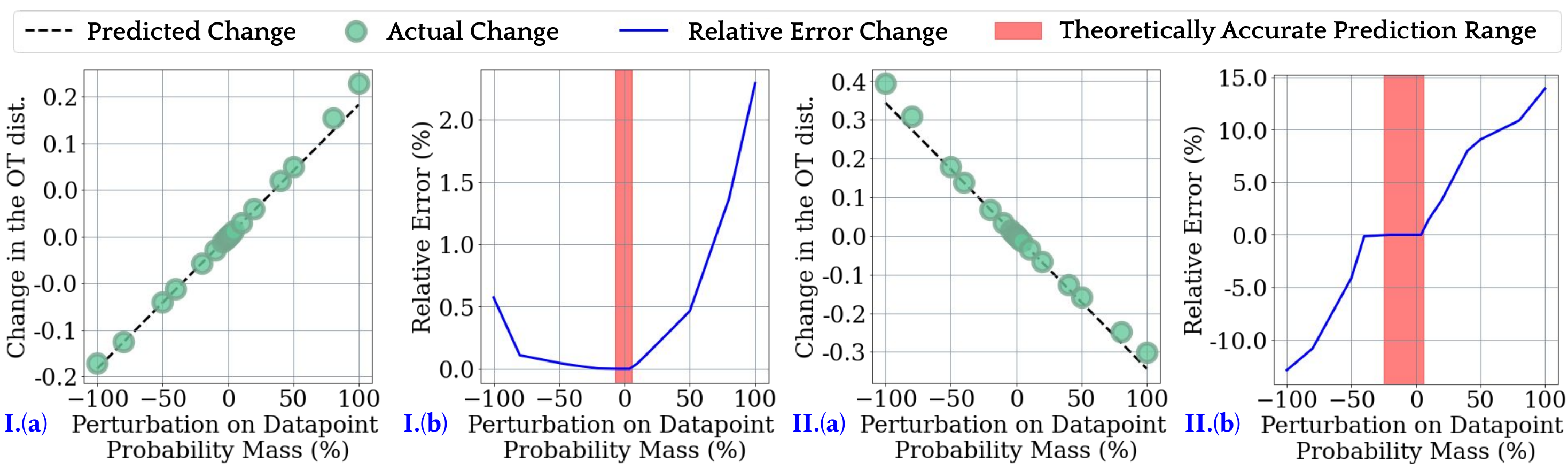}
	\vspace{-1em}
	\caption{Predicting the change to the OT distance for increasing/reducing the probability mass of a point. The OT distance is calculated between two subsets of CIFAR-10. We examine the change to the OT distance predicted by the calibrated gradients against the actual change. The results of two datapoints are visualized for demonstration (\textcolor{blue}{I.(a)/(b)} and \textcolor{blue}{II.(a)/(b)} are analyzed on datapoint \#1 and \#2, respectively). The probability mass of the datapoint is perturbed from $-100\%$ (removing the datapoint) to 100\% (duplicating the datapoint). \textcolor{blue}{I.(a)} and \textcolor{blue}{II.(a)}: Predicted change on the OT distance against the actual change. The predicted change demonstrated high consistency to the actual change despite minor deviation for large perturbation. \textcolor{blue}{I.(b)} and \textcolor{blue}{II.(b)}: Relative error for the prediction, defined as (predicted\_change - actual\_change)/actual\_change$\times100\%$. The color bar represents the theoretical range of perturbation where the change in the OT distance can be accurately predicted. The prediction holds approximately well beyond the range.
    }
    \label{fig:change}
    \vspace{-1em}
\end{figure*}

% In a typical example, 
In our evaluation, we find that this range is about $5\%$ to $25\%$ of the probability measure of the datapoint ($\mu_{(\cdot)}(z_i)$) for perturbations in both directions
% (ranges  $\approx 5\% - 25\%$) 
% of the probability mass of a datapoint
and the pattern seems independent of the size of the datasets. This range being less than the probability mass of a datapoint suggests that we are only able to predict the change to the OT distance for removing/adding a datapoint to the dataset approximately, though, the relative error is well acceptable (depicted in Figure~\ref{fig:change}).

% \textcolor{blue}{[will add a plot here, already have some data points, consist with expections]}
\vspace{-0.5em}
% \end{remark}
\subsection{{Precise recovery of ranking for data values obtained from entropy-OT}}%Quantifying the approximation error in calibrated gradients due to entropy penalization}
Due to computational advantages of the entropy-OT (defined in Eq.~\ref{EOT}), one needs to resort to the solutions to entropy-OT to calculate data values.
% provides significant advantages for its efficient solution algorithms and in large-scale implementations. 
% However, it also results in deviations in the solutions for its approximation nature. 
We quantify the deviation in the calibrated gradients caused by the entropy regularizer. This analysis provides foundations on the potential impact of the deviation on the applications built on these gradients.

\begin{theorem}
\label{thm:approx_bound}
% Let $A$ and $B$ be two datasets each consisting of $m$ and $n$ datapoints, where $A=\{a_1, ... a_m,\}$ and $B=\{b_1, ... b_n\}$, and identical probability mass for all datapoints in a dataset $ \text{prob}(a_i) = 1/m,\,\, \text{prob}(b_i) = 1/n,\,\, \forall i$. 
Let $\operatorname{OT}(\mu_t,\mu_v)$ and $\operatorname{OT_\varepsilon}(\mu_t,\mu_v)$ be the original formulation and entropy penalized formulation (as defined in  \ref{EOT}) for the OT problem between the empirical measures $\mu_t$ and $\mu_v$ associated with the two datasets $\mathbb{D}_t$ and $\mathbb{D}_v$, respectively, where $|\mathbb{D}_t|=N$ and $|\mathbb{D}_v|=M$. Then, for any $i\neq j\neq k \in \{1, 2, \dots, N\}$ and $o\neq p\neq q \in \{1, 2, \dots, M\}$, the difference between the calibrated gradients for two datapoints $z_i$ and $z_k$ in dataset $\mathbb{D}_t$ and the difference for $z_p'$ and $z_q'$ in $\mathbb{D}_v$ can be calculated as
\begin{small}
\begin{equation}
\label{eqn:opt_diff}
\frac{\partial\operatorname{OT}(\mu_t,\mu_v)}{\partial\operatorname{\mu_t}(z_i)}-\frac{\partial\operatorname{OT}(\mu_t,\mu_v)}{\partial\operatorname{\mu_t}(z_k)}=\frac{\partial\operatorname{OT_\varepsilon}(\mu_t,\mu_v)}{\partial\operatorname{\mu_t}(z_i)}-\frac{\partial\operatorname{OT_\varepsilon}(\mu_t,\mu_v)}{\partial\operatorname{\mu_t}(z_k)}-\varepsilon\cdot\frac{N}{N-1}\cdot\left(\frac{1}{(\pi_\varepsilon^*)_{kj}}-\frac{1}{(\pi_\varepsilon^*)_{ij}}\right),
\end{equation}
\begin{equation}
\label{eqn:opt_diff_val}
\frac{\partial\operatorname{OT}(\mu_t,\mu_v)}{\partial\operatorname{\mu_v}(z_p')}-\frac{\partial\operatorname{OT}(\mu_t,\mu_v)}{\partial\operatorname{\mu_v}(z_q')} = \frac{\partial\operatorname{OT_\varepsilon}(\mu_t,\mu_v)}{\partial\operatorname{\mu_v}(z_p')}-\frac{\partial\operatorname{OT_\varepsilon}(\mu_t,\mu_v)}{\partial\operatorname{\mu_v}(z_q')}-\varepsilon\cdot\frac{M}{M-1}\cdot\left(\frac{1}{(\pi_\varepsilon^*)_{\edit{qo}}}-\frac{1}{(\pi_\varepsilon^*)_{\edit{po}}}\right),
\end{equation}
\end{small}

where $\pi_\varepsilon^*$ is the optimal primal solution to the entropy penalized OT problem defined in \ref{EOT}, $z_j$ is any datapoint in $\mathbb{D}_t$ other than $z_i$ or $z_k$, and $z_o'$ is any datapoint in $\mathbb{D}_v$ other than $z_p'$ or $z_q'$.

\end{theorem}

The gradient difference on the left-hand side of (\ref{eqn:opt_diff}) represents the \emph{groundtruth} value difference between two training points $z_i$ and $z_k$ as the values are calculated based on the original OT formulation. In practice, for the sake of efficiency, one only solves the regularized formulation instead and, therefore, this groundtruth difference cannot be obtained directly. Theorem~\ref{thm:approx_bound} nevertheless indicates a very interesting fact that one can calculate the groundtruth difference based on the solutions to the regularized problem, because every term in the right-hand side only depends on the solutions to the regularized problem. Particularly, the groundtruth value difference is equal to the value difference produced by the regularized solutions plus some calibration terms that scale with $\varepsilon$  \citep{nutz2021entropic}. This result indicates that while it is not possible to obtain \emph{individual groundtruth value} by solving the regularized problem, one can actually \emph{exactly} recover the groundtruth value \emph{difference} based on the regularized solutions. In many applications of data valuation such as data selection, it is the order of data values that matters~\citep{kwon2021beta}. For instance, to filter out low-quality data, one would first rank the datapoints based on their values and then throw the points with lowest values. In these applications, solving the entropy-regularized program is an ideal choice---which is both efficient and recovers \edit{the exact ranking of datapoint values}. Finally, note that Eq.~\ref{eqn:opt_diff_val} presents a symmetric result for the calibrated gradients for validation data.
In our experiments, we set $\eps=0.1$, \edit{rendering} the corresponding calibration terms to be negligible. \edit{As a result, we can} directly use the calibrated gradients solved by the regularized program to \edit{rank datapoint values}. (Note that for the calibrated gradients computed by the proposed methods, a positive value indicates increasing the presence of this data will increase the OT distance and decrease expected performance. \textit{We consider the "value" of a datapoint as the negation of calibrated gradients.})

% \begin{proof}\renewcommand{\qedsymbol}{}
% See Appendix 
% \end{proof}
% \begin{remark}[error bound for entropy-OT]
% \textcolor{blue}{say a few words on what this result means}
% \end{remark}

% talk about ways of measure value of individual points and their limitations

% introduce our dual formulation and gradients

% talk about the properties of this data value measure such as fairness

\section{Experiments} \label{expe}
\vspace{-0.5em}
% questions/hypotheses:
% - do our proposed dataset utiloity and data values make sense? (coorespond to quality data)
% - how do different hypers in algorihtm impact the performance?
% We call the proposed data valuation approach based on the calibrated gradients, \AlgName. 
In this section, we demonstrate the practical efficacy and efficiency of $\AlgName$ on various classification datasets. We compare with nine baselines: (1) Influence functions (INF)~\citep{koh2017understanding}, which approximates the LOO error with first-order extrapolation; (2) TracIn-Clean~\citep{pruthi2020estimating}, which accumulates the loss change on validation data during training whenever the training point of interest is sampled; (3) TracIn-Self~\citep{pruthi2020estimating}, which is similar to TracIn-Clean but accumulates the training loss changes; (4) KNN-Shapley (KNN-SV)~\citep{jia2019efficient}, which approximates the Shapley value using K-Nearest-Neighbor as a proxy model; and (5) Random, a setting where we select a random subset from the target dataset. We also consider the popular data valuation approaches: (6) Permutation Sampling-based Shapely value (Perm-SV)~\citep{jia2019towards}, (7) Least Cores (LC)~\citep{yan2021if}, (8) TMC-Shapley (TMC-SV) and (9) G-Shapley (G-SV)~\citep{ghorbani2019data}. Baselines (6)-(9) are, however, computationally infeasible for the scale of data that we study here. So we exclude them from the evaluation of efficacy in different use cases. We also provide a detailed runtime comparison of all baselines. For all methods to be compared, a validation set of $10,000$ samples is assumed. \edit{For our method, we first use the validation data to train a deep neural network model PreActResNet18~\citep{he2016identity} from scratch for feature extraction. Then, from its output, we compute the class-wise Wasserstein distance and the calibrated gradients for data valuation. Details about datasets, models, hyperparameter settings, and ablation studies of the hyperparameters and validation sizes are provided in Appendix~\ref{app:exp}.} %Done \yi{hoang, please add the abbreviations to those methods to match the figure} \doublecheck{add embedder info, the validation size.}

We evaluate on five different use cases of data valuation: \emph{detecting backdoor attack, poisoning attack, noisy features, mislabeled data, and irrelevant data}. The first four are conventional tasks in the literature and the last one is a new case. All of them have a common goal of identifying ``low-quality'' training points. To achieve this goal, we rank datapoints in ascending order of their values and remove some number of points with lowest data values. For each removal budget, we calculate the \emph{detection rate}, i.e., the percentage of the points that are truly bad within the removed points.
% For the use cases that involve attack detection, we will also evaluate the attack 

\begin{figure*}
% \vspace{-2.5em}
    \centering%
	\includegraphics[height=6.5cm, width=0.92\linewidth]{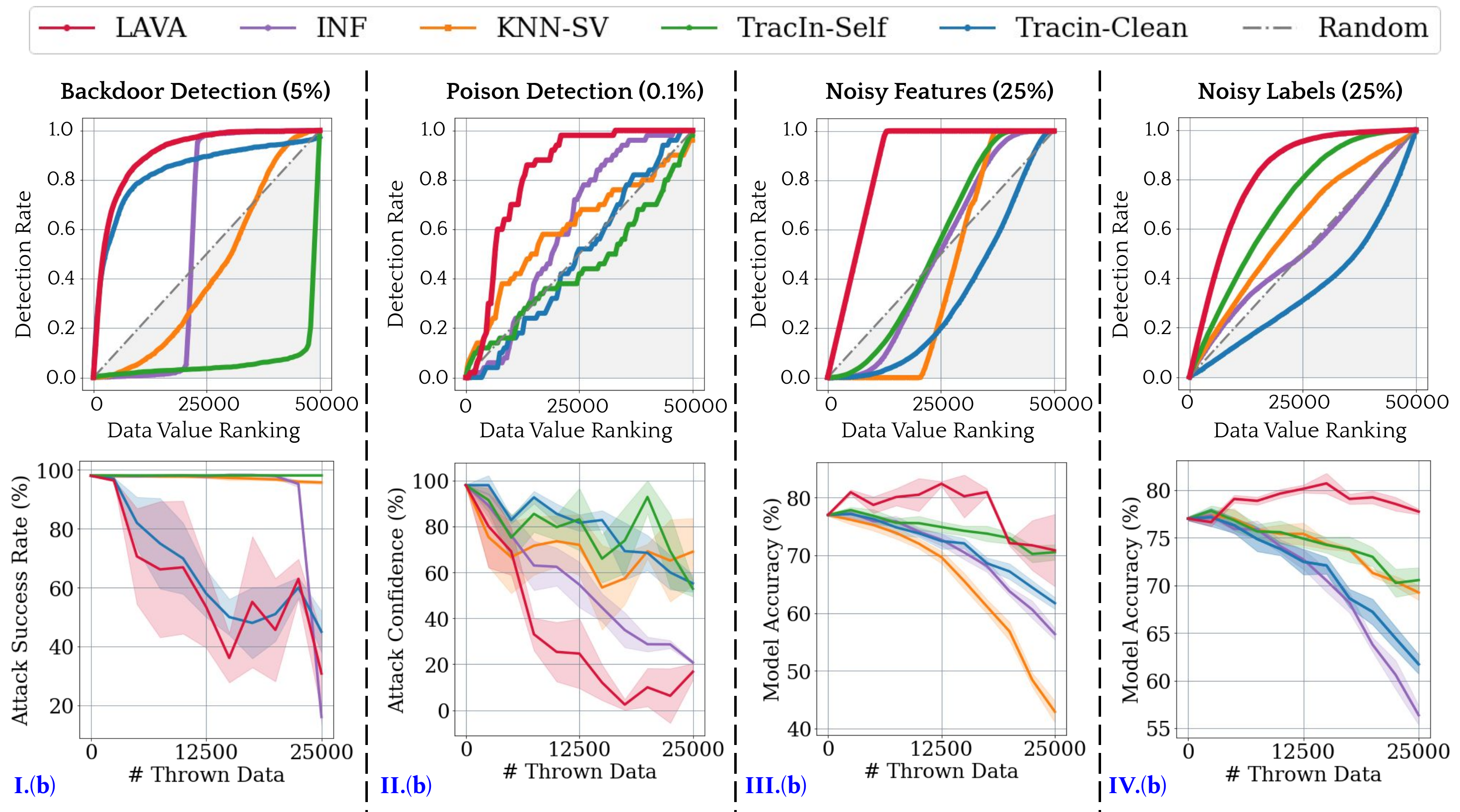}
	\caption{Performance comparison between $\AlgName$ and baselines on various use cases. 
	 For \textcolor{blue}{I.(b)}, we depict the Attack Accuracy, where lower value indicates more effective detection. For \textcolor{blue}{II.(b)}, we depict the Attack Confidence, as lower confidence indicates better poison removal. For \textcolor{blue}{III.(b)} and \textcolor{blue}{IV.(b)}, we show the model test accuracy, where higher accuracy means effective data removal.
% 	\kang{why in Fig. 3(b) that LAVA has a higher model accuracy at 0 thrown data?\hoang{it's actually not 0, but 2500 removed data based on our inspection}}
% 	\yi{I did sth wrong in I(a)/II(a)?}
    }
    \label{fig:all_detection}
\end{figure*}

% #Figure: data removal performance - poisoned, mislabeled, backdoor, feature noise

% Here, we consider a case when the training set was either intentionally or unintentionally modified, which cause the model performance degrade in certain cases. We show the effectiveness of $\AlgName$ and compare the performance with baselines. All prepared samples and attack settings are further provided in the Appendix.

\textbf{Backdoor Attack Detection.} \label{expe_back}
% Backdoor attacks inject maliciously constructed
% data into a training set so that, at test time, the trained
% model misclassifies inputs patched with a backdoor trigger as
% an adversarially-desired target class. In the main text, we consider XXX, a popular backdoor attack algorithm~\citep{}, which injects training points that contain a backdoor trigger and relabeled as a target class. The evaluation on other types of backdoor attacks can be found in Appendix. 
A popular technique of introducing backdoors to models is by injecting maliciously constructed
data into a training set~\citep{zeng2021adversarial}. At test time, any trained model would misclassify inputs patched with a backdoor trigger as
the adversarially-desired target class. In the main text, we consider the Trojan Square attack, a popular attack algorithm~\citep{liu2017trojaning}, which injects training points that contain a backdoor trigger and are relabeled as a target class. The evaluation of other types of backdoor attacks can be found in Appendix~\ref{app:exp}. 
% In the backdoor attack, we perform a Trojan backdoor attack with square patches ("triggers") [X], by poisoning random samples with a certain patch and modify the label to a specific target class to make the model during training associate that patch with the target class. This way, at the inference time, any test example containing the patch will be misclassified into the target class. What makes this attack dangerous is that it is hard to detect, since the training accuracy is almost untouched due to low poisoning rate and so the Trojan attack is deployed unnoticed. 
To simulate this attack, we select the target attack class \textsl{Airplane} and poison $2500\ (5\%)$ samples of the total CIFAR-10 training set ($50k$) with a square trigger. In Figure~\ref{fig:all_detection} \textcolor{blue}{I.(a)}, we compare the detection rates of different data valuation methods.
% show the detection rate of the poisoned images using the inspection order according to our data value metric. 
$\AlgName$ and TracIn-Clean outperform the others by a large margin. In particular, for LAVA, the first $20\%$ of the points that it removes contain at least $80\%$ of the poisoned data.
% and inspecting only the first $20\%$ of all data. 
We also evaluate whether the model trained after the removal still suffers from the backdoor vulnerability. To perform this evaluation, we calculate the \emph{attack accuracy}, i.e., the accuracy of the model trained on the remaining points to predict backdoored examples as the target label. A successful data removal would yield a lower attack accuracy. Figure~\ref{fig:all_detection} \textcolor{blue}{I.(b)} shows that our method already takes effect in the early stages, whereas other baselines can start defending from the attack only after removing over $13,000$ samples.
% shows us how the attack accuracy decreases when we gradually remove data according our data value ranking. Our method already takes effect in early stages, when the Trojan patches samples are removed. Other baselines can start defending from the attack only after removing over $13,000$ samples. However, $\AlgName$ performs well already at early stages, 
The efficacy of $\AlgName$ is in part attributable to inspection of distances between both features and labels. 
The backdoored training samples that are poisoned to the target class will be ``unnatural'' in that class, i.e., they have a large feature distance from the original samples in the target class. While the poisoned examples contain a small feature perturbation compared to the natural examples from some other classes, their label distance to them is large because their labels are altered.
% .e. the in-class feature distances between original samples will deviate more than expected. % Ruoxi: why woult feature diance between clean samples change?
% Additionally, b
% By changing labels to a target class,  
% the class 'airplane' has an inflated number of unnatural samples, which causes the change in the feature distribution of the class. It   This phenomena is well reflected in our method's calculations which explains a higher detection rate than other baselines.

\textbf{Poisoning Attack Detection.} Poisoning attacks are similar to backdoor attacks in the sense that they both inject adversarial points into the training set to manipulate the prediction of certain test examples. However, poisoning attacks are considered unable to control test examples.  
% causing a certain target sample to be mispredicted to some poisoned class during the inference time. 
We consider a popular attack termed ``feature-collision'' attack~\citep{shafahi2018poison}, where we select a target sample from the \textsl{Cat} class test set and blend the selected image with the chosen target class training samples, \textsl{Frog} in our case. In this attack, we do not modify labels and blend the \textsl{Cat} image only into $50$ ($0.1\%$) samples of \textsl{Frog}, which makes this attack especially hard to detect. During inference time, we expect the attacked model to consistently classify the chosen \textsl{Cat} as a \textsl{Frog}. In Figure~\ref{fig:all_detection} \textcolor{blue}{II.(a)}, we observe that $\AlgName$ outperforms all baselines and achieves an $80\%$ detection rate by removing only 11k samples, which is around $60\%$ fewer samples than the highest baseline. Figure~\ref{fig:all_detection} \textcolor{blue}{II.(b)} shows that by removing data according to $\AlgName$ ranking, the target model has reduced the confidence of predicting the target \textsl{Cat} sample as a \textsl{Frog} to below $40\%$. Our technique leverages the fact that the features from a different class are mixed with the features of the poisoned class, which increases the feature distance between the poisoned and non-poisoned \textsl{Frog} examples.
% We consider a popular feature collision-based poisoning attack~\citep{}, where we select a target sample of the Cat class, and blend an image from a chosen target class, Frog in our case. In this attack, we do not modify labels, and blend the 'cat' image only into $50$ ($0.1\%$) samples of 'frog', which makes this attack especially hard to detect. During inference time, we expect the attacked model to consistently classify the chosen 'cat' into a frog. In the Figure~\ref{fig:all_detection} II.(a) we observe that $\AlgName$ outperforms all baselines and achieves $80\%$ detection rate by removing only 11k samples, which is around $60\%$ less samples than the highest baseline. Figure~\ref{fig:all_detection} II.(b) shows that removing data according to $\AlgName$ ranking, the target model has reduces the confidence of predicting the target 'cat' sample as a 'frog' to below $40\%$. Our technique leverages the fact that the features from a different class are mixed with the features of the poisoned class, which increase the feature distance between the poisoned and non-poisoned Frog examples.

\textbf{Noisy Feature Detection.}
While adding small Gaussian noises to training samples may benefit model robustness~\citep{rusak2020simple},
% but unexpectedly can also improve the model's robustness \citet{rusak2020simple}. However, when considering strong noises, which may also appear in a natural setting, the model's
strong noise, such as due to sensor failure, can significantly affect the model performance. We add strong white noise to $25\%$ of all CIFAR-10 dataset without changing any labels. Our method performs extremely well as shown in Figure~\ref{fig:all_detection} \textcolor{blue}{III.(a)} and detects all 12,500 noisy samples by inspecting less than 15,000 samples. This explains the sudden drop of the model's accuracy at the removal budget of 15,000 samples in Figure~\ref{fig:all_detection} \textcolor{blue}{III.(b)}: the model starts throwing away only clean samples from that point. $\AlgName$ performs well in this scenario since the strong noise increases the feature distance significantly.

\begin{wrapfigure}{R}{0.6\columnwidth}
    \vspace{-1em}
    \centering
    \includegraphics[width=0.53\columnwidth]{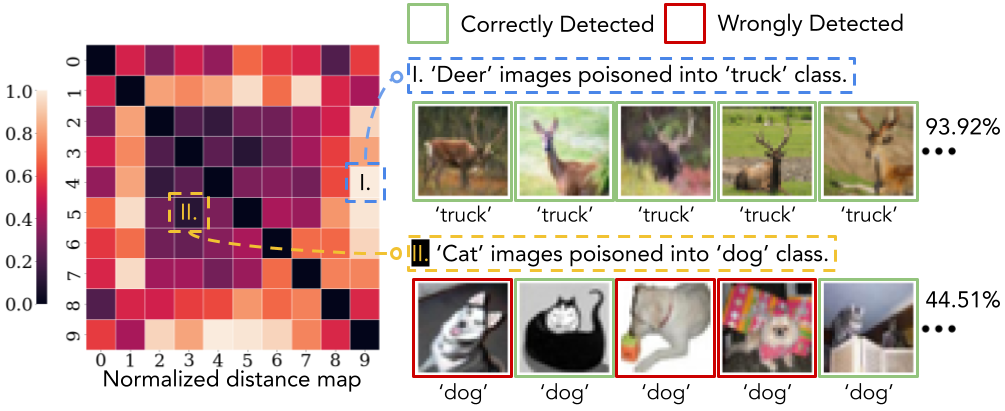}
    \vspace{-1em}
    \caption{Left: Heatmap of inter-class distance of CIFAR-10. Right: Examples of irrelevant data. The detection rates of first 500 inspected images are $94\%$ and $46\%$ for \textsl{Deer}-\textsl{Truck}, \textsl{Cat}-\textsl{Dog}, respectively.
    }
    \vspace{-1em}
    \label{fig:data_irl}
\end{wrapfigure}

\textbf{Mislabeled Data Detection.} \label{sec:mislab}Due to the prevalence of human labeling errors~\citep{karimi2020deep}, it is crucial to detect mislabeled samples. We shuffle labels of $25\%$ samples in the CIFAR-10 dataset to random classes. Unlike backdoor and poisoning attacks, this case is especially harder to detect since wrong samples are spread out throughout classes instead of all placed inside a target class. However, as shown in Figure~\ref{fig:all_detection} \textcolor{blue}{IV.(a)}, $\AlgName$'s detection rate outperforms other baselines and the model performance is maintained even after 20k of removed data (Figure \textcolor{blue}{IV.(b)}).
% When considering features from the original classes, the distance between correctly and incorrectly labeled is increased due to label distance. When considering samples in a class, the argument is flipped to feature distance.

% The result shows that our method can detect poisoned data in earlier stages than other methods, which effectively allows $\AlgName$ to remove bad data much earlier.

%  Figure: data selection performance - data summarization, debiasing [X] Based on Tianhao's resasoining, weh should not reuse the debiasing experiment.

% \begin{figure*}
%     \centering%
% 	\includegraphics[width=0.99\linewidth]{figs/heatmap.png}
% 	\caption{Detecting irrelevant images within each class. a.) Heatmap of class distances based on the same feature extractor.  b.)  Heatmap of class distances based on a different feature extractor.
% 	Detection rates of first 500 inspected images are $94\%, 45\%$, $69\%$ for 'deer'-'truck', 'cat'-'dog', and 'plane'-'bird' poisoning, respectively.}
%     \label{fig:data_irl}
%     % \vspace{-0.5em}
% \end{figure*}

\textbf{Irrelevant Data Detection.} Often the collected datasets through web scraping have irrelevant samples in given classes~\citep{northcutt2021pervasive,tsipras2020imagenet}, e.g.,
% in a class of Tiger, we might encounter images of trees \yi{I think this example is a bit confusing, the later one is good}, or 
in a class of \textsl{Glasses}, we might have both water glass and eyeglasses due to lack of proper inspection or class meaning specification. This case is different from the mislabeled data scenario, in which case the training features are all relevant to the task.
% contain any class of irrelevant samples and thus not train on the irrelevant samples. 
Since the irrelevant examples are highly likely to have completely different features than the desired class representation, $\AlgName$ is expected to detect these examples.
% with calibrated gradients of the OT w.r.t each sample in the class. 
% The images with higher calibrated gradients are more probable to be out of class, since they cause bigger distance increase due to lack of similar samples in the class. On the other hand, the images that are representative of the class should have lower gradient, and removing one of those images will not change the representative of the class much. 
We design an experiment where we remove all images of one specific class from the classification output but split them equally to the other remaining classes as irrelevant images. As shown in Figure~\ref{fig:data_irl}, the detection result over a class varies based on the distance between that class and the class from which irrelevant images are drawn. For instance, when \textsl{Deer} images are placed into the \textsl{Truck} class, we can detect almost $94\%$ of all \textsl{Deer} images within first $500$ removed images. On the other hand, when we place \textsl{Cat} images into \textsl{dog} class, our detection rate drops to $45\%$ within the top $500$.

\begin{wrapfigure}{R}{0.48\columnwidth}
    \centering
    \vspace{-1em}
    \includegraphics[width=0.45\columnwidth]{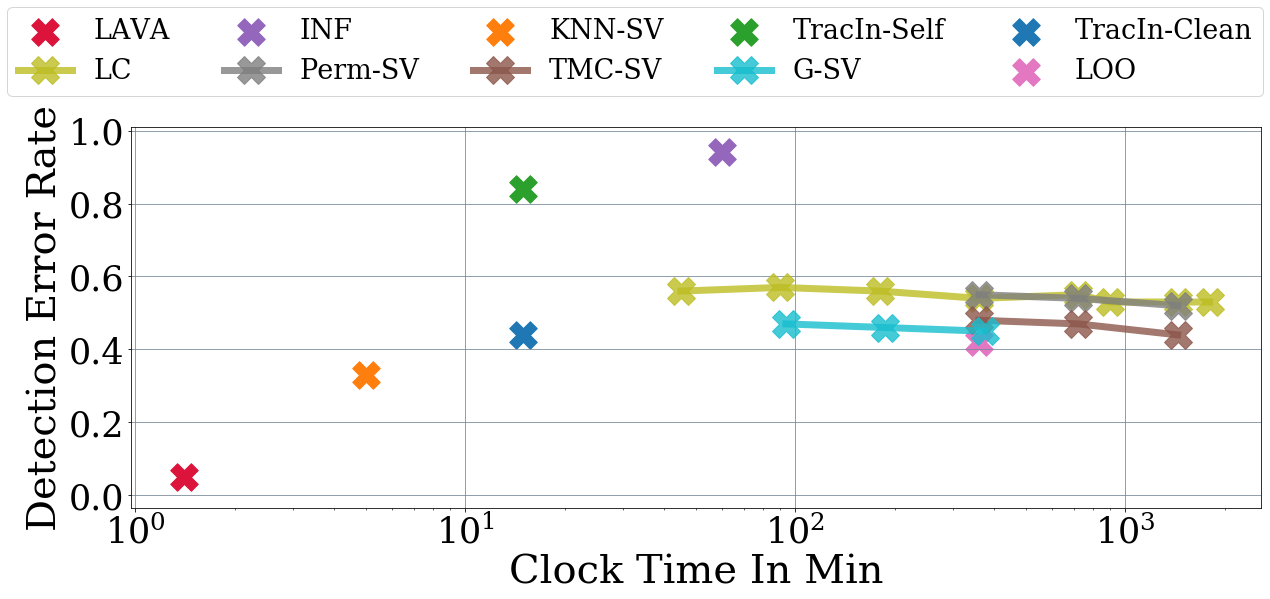}
    \vspace{-1.2em}
    \caption{Runtime v.s. Detection Error comparison between $\AlgName$ and baselines on inspecting 2000 samples from CIFAR-10 with $10\%$ backdoor data.
    }
    \label{fig:runtime}
    \vspace{-0.85em}
\end{wrapfigure}

\textbf{\edit{Computational} Efficiency.} So far, we have focused on the method's performance without considering the actual runtime. We compare the runtime-performance tradeoff on the CIFAR-10 example of $2000$ samples with $10\%$ backdoor data, a scale in which every baseline can be executed in a reasonable time. As shown in Figure~\ref{fig:runtime}, our method achieves a significant improvement in efficiency while being able to detect bad data more effectively.

\edit{\textbf{Dependence on Validation Data Size.}} 
For current experiments, we have assumed the validation set of size $10K$. Such a scale of data is not hard to acquire, as one can get high-quality data from crowdsourcing platforms, such as Amazon Mechanical Turk for $\$12$ per each $1K$ samples \citep{amazon2019machine}. While our method achieves remarkable performance when using $10K$ validation data, we perform ablation study on \edit{much} smaller sets (Appendix~\ref{app:val_size}), where $\AlgName$, notably, can still outperform other baselines. As an example on mislabeled data detection, our method with $2K$ validation data achieves $80\%$ detection rate at data removal budget of $25K$ (Fig.~\ref{fig:misl-lbl-wgt}), whereas the best performing baseline achieves such a performance with $5$ times bigger validation data, $10K$ (Fig.~\ref{fig:all_detection} IV.(a)). Furthermore, even on a tiny validation set of size $500$, $\AlgName$ consistently outperforms all the baselines with the same validation size (Fig.~\ref{fig:valid-size-500}). This shows that our method remains effective performance for various sizes of validation data.

% \doublecheck{replace this with figure.}
% \begin{wraptable}{r}{0cm}
% \begin{tabular}{l|rr}
% \textbf{Method} & \textbf{Time (min)}    \\  % & Detection Error Rate 
% \hline
% TMC-Shapley     & 360                           \\ %  & 0.48 
% LOO             & 195.64                        \\ %  & 0.42
% G-Shapley       & 93.6                         \\ %   & 0.47
% INF             & 60                            \\ %  & 0.96
% LC              & 45                           \\  %  & 0.54 
% Tracin-Self     & 15                             \\ % & 0.84
% Tracin-Clean    & 15                           \\ %   & 0.44
% KNN             & 5                             \\ %  & 0.33
% \textbf{LAVA}   & \textbf{1.4}                     %  & 0.05
% \end{tabular}
% \caption{Runtime comparison between $\AlgName$ and baselines.\yi{add a column showing the efficacy? E.g., detection rate?} \hoang{Actually we have results, the problem is that the case is small for those baselines to even work well, in our case it does not matter. but for other baselines they might need more data to train a model significantly. It's commented out} \yi{re-scale your table to mach the text size of Figure 2}\yi{figure?}}
% \label{table:runtime}
% \end{wraptable}

\vspace{-0.5em}
\section{Related Work} \label{related}
\vspace{-0.5em}
% Due to its substantial impact on real-world applications, quantifying data value has emerged as an essential topic. Many related works have been proposed to measure the suitable value of data to compensate data contributors. 

Existing data valuation methods include LOO and influence function \citep{koh2017understanding}, the Shapley value \citep{jia2019towards,ghorbani2019data,wang2023note}, the Banzhaf value \citep{wang2022data}, Least Cores~\citep{yan2021if}, Beta Shapley \citep{kwon2021beta}, and reinforcement learning-based method \citep{yoon2020data}. However, they all assume the knowledge of the underlying learning algorithms and suffer large computational complexity. The work of~\cite{jia2019efficient} has proposed to use $K$-Nearest Neighbor Classifier as a default proxy model to perform data valuation. While it can be thought of as a learning-agnostic data valuation method, it is not as effective and efficient as our method in distinguishing data quality. 
\cite{xu2021validation} propose to use the volume to measure the utility of a dataset. Volume is agnostic to learning algorithms and easy to calculate because is defined simply as the square root of the trace of feature matrix inner product. However, the sole dependence on features makes it incapable of detecting bad data caused by labeling errors. Moreover, to evaluate the contribution of individual points, the authors propose to resort to the Shapley value, which would still be expensive for large datasets.

\vspace{-0.5em}
\section{Discussion and Outlook} \label{discus}
\vspace{-0.5em}
% This paper describes a learning-algorithm-agnostic data valuation framework. In particular, in contrast to existing methods which typically adopt model validation performance as the data utility function, we approximate the utility of a dataset based on its Wasserstein distance to a given validation set and provide theoretical justification for this approximation. Furthermore, we propose to use the calibrated gradients of the Wasserstein distance to value individual datapoints, which can be obtained for free if one uses off-the-shelf solver to calculate the Wasserstein distance. Importantly, we have tested on various datasets and our framework can greatly improve the state-of-the-art performance of using data valuation methods to detect bad data, while being significantly more efficient. Due to the stochasticity of ML as well as the inherent tolerance to noise, it is often difficult to identify low-quality data by inspecting their influence on model performance scores. The take-away from our empirical study is that despite extensive methods adopted in the past, identifying low-quality data through model performance changes is actually \emph{suboptimal}; lifting the dependence of data valuation on the actual learning process provides a better pathway to distinguish data quality.

This paper describes a learning-agnostic data valuation framework. In particular, in contrast to existing methods which typically adopt model validation performance as the utility function, we approximate the utility of a dataset based on its \edit{class-wise} Wasserstein distance to a given validation set and provide theoretical justification for this approximation. Furthermore, we propose to use the calibrated gradients of the \edit{OT} distance to value individual datapoints, which can be obtained for free if one uses an off-the-shelf solver to calculate the Wasserstein distance. Importantly, we have tested on various datasets, and our \edit{$\AlgName$} framework can significantly improve the state-of-the-art performance of using data valuation methods to detect bad data while being substantially more efficient. Due to the stochasticity of ML and the inherent tolerance to noise, it is often challenging to identify low-quality data by inspecting their influence on model performance scores. The take-away from our empirical study is that despite being extensively adopted in the past, low-quality data detection through model performance changes is actually \emph{suboptimal}; lifting the dependence of data valuation on the actual learning process provides a better pathway to distinguish data quality.

Despite the performance and efficiency improvement, our work still has some limitations. As a result, it opens up many new investigation venues: \textbf{(1)} How to further lift the dependence on validation data?
While a validation set representative of the downstream learning task is a common assumption in the ML literature, it may or may not be available during data exchange. 
\textbf{(2)}
Our design could be vulnerable to existing poisons that directly or indirectly minimize the similarity to clean data~\citep{huang2021unlearnable,zeng2022narcissus}. Further investigation into robust data valuation would be intriguing.
\textbf{(3)} Our current method does not have enough flexibility for tasks that aim for goals beyond accuracy, e.g., fairness. Folding other learning goals in is an exciting direction.
\textbf{(4)} Customizing the framework to natural language data is also of practical interest.

\section{Acknowledgements} \label{ack}

RJ and the ReDS Lab gratefully acknowledge the support from the Cisco Research Award, the Virginia Tech COE Fellowship, and the NSF CAREER Award. Jiachen T. Wang is supported by Princeton’s Gordon Y. S. Wu Fellowship. YZ is supported by the Amazon Fellowship.

\bibliography{iclr2023_conference.bbl}
\bibliographystyle{iclr2023_conference}

% \kang{Editorial checklist:\\\\
% \sout{a. unify the notation of "OT" and "W";\\
% b. update the theorem and proof in Appendix;\\
% c. (everyone) checks all the edited text (marked in blue) and remove the marks;}\\
% d. read through every sentence in the manuscript, checking for insensible expressions/typos/grammar mistakes/inconsistencies/fallacies;\\
% \sout{e. make sure the structure and highlights of the paper are clear enough and allow no space for misinterpretation;}\\
% f. add the list for conflicts of interests.}
% \ruoxi{\\ \sout{g. Go through the review comments from last cycle and revise the paper to address some common concerns, like how to acquire validation data, etc.}\\}
% \kang{\sout{h. main text limit: 9 pages strict.}\\
% i. Equation indexes are dropped, check Eq references for consistency}

\newpage

\appendix{}
% \rule{\hsize}{4pt} %[0.25\baselineskip] 
% \begin{center}
% \LARGE{\textbf{Appendix -- ``Data Valuation without Pre-Specified Learning
% Algorithms''\\\kang{Is this the correct format?}}} \\[0.75\baselineskip]
% \large{\textbf{Appendix}}
%\end{center}
% \rule{\hsize}{2pt} 

\normalsize
\begin{appendices}
\section{Restatement of Theorems and Full Proofs}
\label{appendix:proof-bound}
In this section, we will restate our main results and give full proofs.

\subsubsection{\add{Summary of Notations}}
\add{Let $\mut$, $\muv$ be the training and validation distribution, respectively. We denote $\ft: \X \rightarrow \zeroandone$, $\fv: \X \rightarrow \zeroandone$ as the labeling functions for training and validation data. 
We can then denote the \emph{\textbf{joint distribution of random data-label pairs}} $(x,\ft(x))_{x \sim \mut(x)}$ and $(x, \fv(x))_{x \sim \muv(x)}$ \emph{\textbf{as $\mutft$ and $\muvfv$, respectively, which are the same notations as $\mu_t$ and $\mu_v$ but made with explicit dependence on $f_t$ and $f_v$ for clarity}}. The distributions of $(\fv(x))_{x \sim \muv(x)}$, $(\ft(x))_{x \sim \mut(x)}$ are denoted as $\mufv, \muft$, respectively. 
Besides, we define conditional distributions
$\mut(x|y) := \frac{ \mut(x)I[\ft(x)=y] }{ \int \mut(x)I[\ft(x)=y] dx }$ and $\muv(x|y) := \frac{ \muv(x)I[\fv(x)=y] }{ \int \muv(x) I[\fv(x)=y] dx }$. 
Let $\model: \X \rightarrow \zeroone$ be the model trained on training data, and $\mL: \zeroone \times \zeroone \rightarrow \R^+$ be the loss function. 
We denote $\pi \in \Pi(\mu_1, \mu_2)$ as a coupling between a pair of distributions $\mu_1, \mu_2$, and $\metric: \X \times \X \rightarrow \R$ as a distance metric function. }

\add{
The 1-Wasserstein distance with respect to distance function $\metric$ between two distributions $\mu_1, \mu_2$ is defined as $W_{\metric}(\mu_1, \mu_2) := \inf_{\pi \in \Pi(\mu_1, \mu_2)} \underset{(x, y) \sim \pi}{\E}\left[\metric(x, y)\right]$. 
More generally, the 1-Wasserstein distance with respect to cost function $\C$ is defined as  $W_{\C}(\mu_1, \mu_2) := \inf_{\pi \in \Pi(\mu_1, \mu_2)} \underset{(x, y) \sim \pi}{\E}\left[ \C(x, y) \right]$. }

\subsubsection{\add{Statement of Assumptions}}

\add{To prove Theorem \ref{thm:bound}, we need the concept} of probabilistic \emph{cross}-Lipschitzness, which assumes that two labeling functions should produce consistent labels with high probability on two close instances. 
\begin{definition}[Probabilistic Cross-Lipschitzness]
%Let $\mu_t$ and $\mu_v$ be respectively the training and validation distributions. 
Two labeling functions $f_t:\X\rightarrow \add{\{0,1\}}$ and $f_v:\X\rightarrow \add{\{0,1\}}$ are $(\eps,\delta)$-probabilistic cross-Lipschitz w.r.t. a joint distribution $\pi$ over \add{$\X \times \X$} if for all \add{$\eps>0$}: 
\begin{align}
 P_{(x_1,x_2)\sim \pi}[|f_t(x_1)-f_v(x_2)|>\eps \metric(x_1,x_2)]\leq \delta.
 \end{align}
\end{definition}

\add{Intuitively, given labeling functions $f_v$, $f_t$ and a coupling $\pi$, it bounds the probability of finding pairs of training and validation instances labelled differently in a $(1/\eps)$-ball with respect to $\pi$.} 

% The standard deterministic Lipschitz condition is stated for real-valued functions, namely, for all $x_1,x_2$, $|f(x_1)-f(x_2)|\leq \epsilon d(x_1,x_2)$ for some constant $\epsilon$. This condition can be readily applied to any model outputting the probability of a class. However, when we consider the groundtruth labeling function associated with the training and validation data, whose co-domain is actually constrained to binary values (i.e., $\{0,1\}$), the Lipschitz constant $\epsilon$ would impose a gap of $1/\epsilon$ between points of different classes. Thus, it requires the domain to be disconnected or the label function to be constant, the requirement which might not be practical. The probabilistic Lipschitzness condition is a relaxation of the standard Lipschitz condition and prescribes that the standard Lipschitz condition is satisfied with high probability. This
% notion of probabilistic Lipschitzness penalizes having high probability
% density in areas that are label heterogeneous.

% \begin{definition}[Probabilistic Lipschitzness (adapted from~\cite{courty2017joint})]
% A function $f: \X \rightarrow \zeroone$ is $(\eps,\delta)$-probabilistic Lipschitz in distance metric $\metric$ w.r.t. a joint distribution $\pi$ over $\X \times \X$ if for all $\eps>0$ we have 
% $$
% P_{(x_1,x_2) \sim \pi} \left[|f(x_1)-f(x_2)|>\eps \metric(x_1,x_2)\right] \leq \delta
% $$
% \kang{Should we clarify on why the labeling function is a mapping to $R$ rather than \{0, 1\}?}\tianhao{I removed the wording ``labeling''.}
% \end{definition}

\textbf{\add{Our Assumptions.}} 
\add{Assuming the model $\model$ is an $\eps$-Lipschitz function in both inputs.} 
\add{Given a metric function $\metric(\cdot, \cdot)$, we define a cost function $\C$ between $(x_t, y_t)$ and $(x_v, y_v)$ as 
}
\begin{align}\add{
    \C( (x_t, y_t), (x_v, y_v)) := \metric(\xt, \xv) + c W_{\metric}( \mut(\cdot | y_t), \muv(\cdot | y_v)  ),}
\end{align}
\add{
where $c$ is a constant. Let $\pixy$ be the coupling between $\mutft, \muvfv$ such that}

\begin{align}
\add{\pixy := \arginf_{\pi \in \Pi (\mutft, \muvfv )} \E_{((\xt, \yt), (\xv, \yv)) \sim \pi} [\C((\xt, \yt), (\xv, \yv))].}
\end{align}

\add{
We define two couplings $\pistar$ and $\pistartil$ between $\mut(x), \muv(x) $ as follows:}

\begin{align}\add{
    \pistar( \xt, \xv) := \int_{\Y} \int_{\Y} \pixy( (\xt, \yt), (\xv, \yv)) \,d\yt d\yv .}
\end{align}

\add{
For $\pistartil$, we first need to define a coupling between $ \muft, \mufv$:
}

\begin{align}
\add{
    \piy(\yt, \yv) := \int_{\X}\int_{\X} \pixy((\xt, \yt), (\xv, \yv)) \,d\xt d \xv 
    }
\end{align}
\add{
and another coupling between $\mutft, \muvfv $:}
\begin{align}
\add{
    \pixytil( (\xt, \yt), (\xv, \yv) ) 
    := \piy(\yt, \yv ) \mut(\xt|\yt) \muv(\xv|\yv) .
    }
\end{align}
\add{
Finally, $\pistartil$ is constructed as follows:
}
\begin{align}
\add{
    \pistartil(\xt, \xv) := 
    \int_{\Y} \int_{\Y} \piy(\yt, \yv) \mut(\xt|\yt)\muv(\xv|\yv)  \,d\yt d \yv.
    }
\end{align}
\add{
It is easy to see that all joint distributions defined above are couplings between the corresponding distribution pairs.
}

We assume that $\ft$, $\fv$ are $(\eps_{tv}, \delta_{tv})$-probabilistic cross-Lipschitz with respect to \add{$\pistartil$} in metric $\metric$.
%All metrics are $\ell_2$ norm\tianhao{[I don't think KR duality generalize to arbitrary metric]} \ruoxi{double check} and $\eps_{tv} \le c$. 
 Additionally, we assume that $\eps_{tv} / \eps \le c$, there exists constant $M$ such that $|\model(x)|, |\ft(x)|, |\fv(x)| \le M$ for all $x \in \X$, and the loss function $\mL$ is $k$-Lipschitz in both inputs.

The assumption of probabilistic cross-Lipschitzness would be violated only when the underlying coupling assigns large probability to pairs of training-validation features that are close enough (within $1/\eps_{tv}$-ball) but labeled differently. However, $\pistartil$ is generally not such a coupling.
Note that $\pistar$ is the optimal coupling between training and validation distributions that minimizes a cost function $\C$ pertaining to both feature and label space. 
Hence, $\piy(\yt, \yv)$, the marginal distribution of $\pistar$ over the training and validation label space, tends to assign high probability to those label pairs that agree. 
On the other hand, $\pixytil$ can be thought of as a coupling that first generates training-validation labels from $\piy$ and then generates the features in each dataset conditioning on the corresponding labels. Hence, the marginal distribution $\pistartil$ of training-validation feature pairs generated by $\pixytil$ would assign high likelihood to those features with the same labels. So, \edit{conceptually}, the probabilistic cross-Lipschitzness assumption should be easily satisfied by $\pistartil$.

\subsubsection{\add{Detailed Proof}}

\begin{customthm}{1}[restated] 
\add{Given the above assumptions, we have }
\begin{align}
    \E_{\add{x \sim \muv(x)}} \left[ \add{\mL}(\add{\fv(x)}, \model(x)) \right]
    \le \E_{\add{x \sim \mut(x)}} \left[ \add{\mL}(\add{\ft(x)}, \model(x)) \right] + \add{k \eps }W_{\add{\C}}(  \add{\mutft}, \add{\muvfv} ) + 2kM \delta_{tv}.
\end{align}
\end{customthm}
\begin{proof}

\begin{align}
    &\E_{\add{x \sim \muv(x)}} \add{\left[  \mL\nadd{(}\fv(x) \nadd{, \model(x))} \right]} \\
    &= \E_{\add{x \sim \muv(x)}} \add{\left[ \mL\nadd{(}\fv(x) \nadd{, \model(x))} \right]} - \E_{\add{x \sim \mut(x)}} \add{\left[ \mL\nadd{(}\ft(x) \nadd{, \model(x))} \right]} + \E_{\add{x \sim \mut(x)}} \add{\left[ \mL\nadd{(}\ft(x) \nadd{, \model(x))} \right]} \\
    &\le \E_{\add{x \sim \mut(x)}} \add{\left[ \mL\nadd{(}\ft(x) \nadd{, \model(x))} \right]} + \left| \E_{\add{x \sim \muv(x)}} \add{\left[ \mL\nadd{(}\fv(x) \nadd{, \model(x))} \right]} - \E_{\add{x \sim \mut(x)}} \add{\left[ \mL\nadd{(}\ft(x) \nadd{, \model(x))} \right]} \right|.
\end{align}

We bound $\left| \E_{\add{x \sim \muv(x)}} \left[ \add{\mL}(\add{\fv(x)}, \model(x)) \right] - \E_{\add{x \sim \mut(x)}} \left[ \add{\mL}(\add{\ft(x)}, \model(x)) \right] \right|$ as follows:

\begin{small}
\begin{align}
    &\left| \E_{\add{x \sim \muv(x)}} \left[ \add{\mL}(\fv(x), \model(x)) \right] - \E_{\add{x \sim \mut(x)}} \left[ \add{\mL}(\ft(x), \model(x)) \right] \right| \\
    &= \left| \int_{\add{\X}^2} \left[ \add{\mL}(\fv(x_v), \model(x_v)) - \add{\mL}(\ft(x_t), \model(x_t)) \right] d\pistar \add{(x_t, x_v)} \right| \\
    &= \left| \int_{\add{\X}^2} \left[ 
    \add{\mL}(\fv(x_v), \model(x_v)) - \add{\mL}(\fv(x_v), \model(x_t)) + \add{\mL}(\fv(x_v), \model(x_t)) - \add{\mL}(\ft(x_t), \model(x_t)) \right] 
    d\pistar\add{(x_t, x_v)} \right| \\
    &\le \underbrace{\int_{\add{\X}^2} \left| 
    \add{\mL}(\fv(x_v), \model(x_v)) - \add{\mL}(\fv(x_v), \model(x_t)) \right| d\pistar\add{(x_t, x_v)}}_{U_1} \\
    &~~~~~+ \underbrace{\int_{\add{\X}^2} \left|
    \add{\mL}(\fv(x_v), \model(x_t)) - \add{\mL}(\ft(x_t), \model(x_t)) \right| d\pistar\add{(x_t, x_v)}}_{U_2},
\end{align}
\end{small}

where the \add{last} inequality is due to triangle inequality. 

\add{N}ow, we bound $U_1$ and $U_2$ separately. \add{For $U_1$, we have}
\begin{align}
    U_1 
    &\le k \int_{\add{\X}^2} |\model(x_v) - \model(x_t)| \,d\pistar\add{(x_t, x_v)} \\
    &\le k\eps \int_{\add{\X}^2} \metric(\xt, \xv) \,d\pistar\add{(x_t, x_v)}, %+ 2kM\delta
\end{align}
\add{where both inequalities are due to Lipschitzness of $\mL$ and $\model$.} 

%It is slightly more involved to bound $U_2$. 
\add{
In order to bound $U_2$, we first recall that $\piy(\yt, \yv) = \int_{\X}\int_{\X} \pixy((\xt, \yt), (\xv, \yv)) \,d \xv d\xv$ and $\pixytil( (\xt, \yt), (\xv, \yv) ) := \piy(\yt, \yv) \mut(\xt|\yt) \muv(\xv|\yv) $, and note that $\piy(\yt, \yv) = \int_{\X}\int_{\X} \pixytil( (\xt, \yt), (\xv, \yv)) \,d \xt d\xv$.}

\add{
Observe that
}
\begin{align}
    U_2 
    \,&\add{= \int_{\X^2} \int_{\Y^2} \left| 
    \mL(\fv(x_v), \model(x_t)) - \mL(\ft(x_t), \model(x_t)) \right| d \pixy( (\xt, \yt), (\xv, \yv) ) }\\
    &\add{= \int_{\Y^2} \int_{\X^2} \left| 
    \mL(\yv, \model(x_t)) - \mL(\yt, \model(x_t)) \right| d \pixy(  (\xt, \yt) ), (\xv, \yv) }\\
    &\add{\le k \int_{\Y^2} \int_{\X^2} \left| 
    \yv - \yt \right| d \pixy( (\xt, \yt), (\xv, \yv) ) } \\
    &= k \int_{\Y^2} \left| 
    y_v - y_t \right| d \add{\piy}(\yt, \yv),
\end{align}
\add{
where the second equality is due to a condition that if $\yt \ne \ft(\xt)$ or $\yv \ne \fv(\xv)$, then $\pixy((\xt, \yt), (\xv, \yv))=0$.
}

% Moreover, we construct another coupling between $\muvfv , \mutft$ as follows:
% \begin{align}
%     \pixytil((\xv, \yv), (\xt, \yt)) = \piy(\yv, \yt) \muv(\xv|\yv) \mut(\xt|\yt)
% \end{align}
% \tianhao{[Lemma: show that it is a coupling between $\muvfv , \mutft$.]}
% and another coupling between $\muv, \mut$ as 
% \begin{align}
%     \pi'(\xv, \xt) = 
%     \int_{\Y} \int_{\Y} \piy(\yv, \yt) \muv(\xv|\yv) \mut(\xt|\yt) d \yv d\yt
% \end{align}
% \tianhao{[Lemma: show that it is a coupling between $\muv , \mut$.]}

\add{
Now we bound $U_2$ as follows:
}
\begin{align}
    U_2 
    &\le k \int_{\Y^2} \left| 
    y_v - y_t \right| d\add{\piy}(\yt, \yv) \\
    &= k \int_{\add{\X^2}} \int_{\add{\Y^2}} \left| 
    y_v - y_t \right|
    d \add{\pixytil( (\xt, \yt), (\xv, \yv))} \\
    &= k \int_{\yv > \yt} \int_{\X^2} 
    \fv(\xv) - \ft(x_t)
    \,d \add{\pixytil((\xt, \yt), (\xv, \yv) )}
    \\
    &~~~~~~+ k \int_{\yv \le \yt} \int_{\X^2} 
    \ft(x_t) - \fv(\xv)
    \,d \add{\pixytil( (\xt, \yt), (\xv, \yv))}, 
\end{align}
where \add{the last step holds since if $\yt \ne \ft(\xt)$ or $\yv \ne \fv(\xv)$ then $\pixytil( (\xt, \yt), (\xv, \yv))=0$}. 

\add{
Define the region $A = \{ (\xt, \xv): f_v(\xv) - f_t(x_t) < \eps_{tv} \metric(\xt, \xv) \}$.
}

\begin{small}
\begin{align}
    & k \int_{y_v > y_t} \int_{\X^2} f_v(\xv)-f_t(x_t) \,d \add{\pixytil( (x_t, y_t), (\xv, y_v)  ) }\\
    &= k \int_{y_v > y_t} \add{\int_{\X^2 \setminus A} \nadd{f_v(\xv)-f_t(x_t) \,d}\pixytil(  (x_t, y_t), (\xv, y_v) )}\\ 
    &~~~~~\add{+}\,k \int_{y_v > y_t} \add{\int_{A} \nadd{f_v(\xv)-f_t(x_t) \,d}\pixytil(  (x_t, y_t), (\xv, y_v) ) }\\
    &\le k \int_{y_v > y_t} \int_{\add{\X^2 \setminus A}} 2M \,d \add{\pixytil(  (x_t, y_t), (\xv, y_v) )} 
    \\ &~~~~~+ k \int_{y_v > y_t} \int_{\add{A}} f_v(\xv)-f_t(x_t) \,d \add{\pixytil(  (x_t, y_t), (\xv, y_v) )}.
\end{align}
\end{small}

\add{
Let's define} $\tilde{f}_t(x_t) = f_t(x_t)$ and $\tilde{f}_v(x_v) = f_v(x_v)$ if $(x_t, x_v) \in A$, 
and $\tilde{f}_t(x_t) = \tilde{f}_v(x_v) = 0$ otherwise (note that $\tilde{f}_v(x_v)-\tilde{f}_t(x_t)\leq \eps_{tv} \metric(\xt, \xv)$ for all \add{$(x_t, x_v) \in \X^2$)}, \add{then we can bound the second term as follows:
}
\begin{align}
    &k \int_{y_v > y_t} \int_{\add{A}} f_v(x_v)-f_t(x_t) \,d\add{\pixytil( (x_t, y_t), (x_v, y_v)  )} \\
    &\add{\le}\,k \int_{y_v > y_t} d\add{\piy}(\yt, \yv) 
    \int_{\add{A}} f_v(x_v)-f_t(x_t) \,d\add{\mut}(\xt|\yt) d\add{\muv}(\xv|\yv)  \\
    &\add{=}\,k \int_{y_v > y_t} d\add{\piy}(\yt, \yv) 
    \int_{\X^2} \tilde{f}_v(x_v) - \tilde{f}_t(x_t) \, d\add{\mut}(\xt|\yt) d\add{\muv}(\xv|\yv) \label{eqn:redefine} \\
    &= k \int_{y_v > y_t} d\add{\piy}(\yt, \yv) \left[
    \add{\E_{\xv \sim \muv(\cdot|\yv)} [ \tilde{f}_v(x_v) ] - \E_{\xt \sim \muv(\cdot|\yt)} [ \tilde{f}_t(x_t) ]} \right] \\
    &\le k \add{\eps_{tv}} \int_{y_v > y_t} d\add{\piy}(\yt, \yv)
    W_{\metric}(
    \add{\mut}(\add{\cdot}|\yt),
    \add{\muv}(\add{\cdot}|\yv))  . \label{eqn:kp}
\end{align}
Inequality (\ref{eqn:kp}) is a consequence of the duality form of the Kantorovich-Rubinstein theorem \add{(\cite{villani2021topics}, Chapter 1).}

Similarly, we have 
\begin{align}
    & k \int_{y_v \le y_t} \int_{\X^2} f_{\add{t}}(x_{\add{v}})-f_{\add{v}}(x_{\add{t}}) \,d\add{\pixytil(  (x_t, y_t), (\xv, y_v) )} \\
    &\le k \int_{y_{\add{v}} \le y_{\add{t}}} \add{\int_{\X^2 \setminus A}} 2M \,d\add{\pixytil(  (x_t, y_t), (\xv, y_v) ) }
    \\ &+ k \add{\eps_{tv}} \int_{y_{\add{v}} \le y_{\add{t}}} d\add{\piy}(\yt, \yv)
    W_{\metric}(  \add{\mut}(\add{\cdot}|\yt), \add{\muv}(\add{\cdot}|\yv) ) .
\end{align}
\add{
Combining two parts, we have
}
\begin{align}
    U_2 
    &\, \add{\le k \int_{\Y^2} \int_{\X^2 \setminus A} 2M \, d\pixytil(  (x_t, y_t), (\xv, y_v) ) } \\
    &\add{+ \, k \eps_{tv} \int_{ \Y^2 } d\piy(\yt, \yv)
    W_{\metric}(  \mut(\cdot|\yt), \muv(\cdot|\yv)  )}\\
    &\le 2kM \delta_{tv} + k \eps_{tv} \int_{ \Y^2 } d\add{\piy}(\yt, \yv)
    W_{\metric}(
    \add{\mut}(\add{\cdot}|\yt),
    \add{\muv}(\add{\cdot}|\yv)  ),
\end{align}
\add{
where the last step is due to the probabilistic cross-Lipschitzness of $\ft, \fv$ with respect to $\pixytil$. }

% We note that a coupling $\pi \in \Pi(\muv, \mut)$ naturally induce a coupling $\pixy \in \Pi( \muvfv , \mutft )$ by
% \begin{align}
%     \pixy((\xv, \yv), (\xt, \yt))
%     = \pi(\xv, \xt) I[ \yv = \fv(\xv) \wedge \yt = \ft(\xt) ]
% \end{align}

Now, combining the bound for $U_1$ and $U_2$, we have
\begin{align}
    &\left| \E_{\add{x \sim \muv(x)}} \add{\left[ \mL\nadd{(}\fv(x)\nadd{, \model(x))} \right]} - \E_{\add{x \sim \mut(x)}} \add{\left[ \mL\nadd{(}\ft(x)\nadd{, \model(x))} \right]} \right| \\
    &\leq  k\eps \int_{\add{\X^2}} \metric(\xt, \xv) d\add{\pi}(x_t, x_v)
     + 2kM \delta_{\add{tv}} + k \add{\eps_{tv}} \int_{ \Y^2 } d\add{\piy}(\yt, \yv)
    W_{\metric}( \add{\mut}(\add{\cdot}|\yt), \add{\muv}(\add{\cdot}|\yv)  )
    \\
    & = k \int_{(\X \times \Y)^2}
    \left[
    \eps \metric(\xt, \xv) + \eps_{tv}W_{\metric}( \add{\mut}(\add{\cdot}|\yt), \add{\muv}(\add{\cdot}|\yv)  )
    \right]
    d\pi^*_{\add{x,y}} ((x_t,y_t), (x_v,y_v)) + 2kM \delta_{tv} \\
    &\le k \int_{(\X \times \Y)^2}
    \left[
    \eps \metric(\xt, \xv) + \add{c \eps_{tv} } W_{\metric}( \add{\mut}(\add{\cdot}|\yt), \add{\muv}(\add{\cdot}|\yv)  )
    \right]
    d\pi^*_{\add{x,y}} ((x_t,y_t), (x_v,y_v)) + 2kM \delta_{tv} \\
    &\add{\,= k \eps \E_{\pixy} \left[ \C( (x_t,y_t), (x_v,y_v) ) \right] + 2kM \delta_{tv}} \\
    &\add{\,= k \eps W_{\C}( \mutft, \muvfv ) + 2kM \delta_{tv},}
\end{align}
\add{
where the last step is due to the definition of $\pixy$. 
This leads to the final conclusion.}
\end{proof}
% \begin{proposition}[Extensions to debiased computational paradigms based on Sinkhorn divergence]
% \begin{equation*}
% \operatorname{S}_\varepsilon(\mu_t,\mu_v)=\operatorname{OT}\eps
% \end{equation*}
% \end{proposition}

\begin{customthm}{5}[restated]
 
Let $\operatorname{OT}(\mu_t,\mu_v)$ and $\operatorname{OT_\varepsilon}(\mu_t,\mu_v)$ be the original formulation and entropy penalized formulation (as defined in Subsection \ref{eq:ot}) for the OT problem between the empirical measures $\mu_t$ and $\mu_v$ associated with the two data sets $D_t$ and $D_v$, respectively. Then, for any $i\neq j\neq k \in \{1, 2, ... N\}$ and $o\neq p\neq q \in \{1, 2, ... M\}$, the difference between the calibrated gradients for two datapoints $z_i$ and $z_k$ in dataset $D_t$ and the difference for $z_p'$ and $z_q'$ in $D_v$ can be calculated as
\begin{small}
\begin{equation*}
\frac{\partial\operatorname{OT}(\mu_t,\mu_v)}{\partial\operatorname{\mu_t}(z_i)}-\frac{\partial\operatorname{OT}(\mu_t,\mu_v)}{\partial\operatorname{\mu_t}(z_k)}=\frac{\partial\operatorname{OT_\varepsilon}(\mu_t,\mu_v)}{\partial\operatorname{\mu_t}(z_i)}-\frac{\partial\operatorname{OT_\varepsilon}(\mu_t,\mu_v)}{\partial\operatorname{\mu_t}(z_k)}-\varepsilon\cdot\frac{N}{N-1}\cdot\left(\frac{1}{(\pi_\varepsilon^*)_{kj}}-\frac{1}{(\pi_\varepsilon^*)_{ij}}\right),
\end{equation*}
\begin{equation*}
\frac{\partial\operatorname{OT}(\mu_t,\mu_v)}{\partial\operatorname{\mu_v}(z_p')}-\frac{\partial\operatorname{OT}(\mu_t,\mu_v)}{\partial\operatorname{\mu_v}(z_q')} = \frac{\partial\operatorname{OT_\varepsilon}(\mu_t,\mu_v)}{\partial\operatorname{\mu_v}(z_p')}-\frac{\partial\operatorname{OT_\varepsilon}(\mu_t,\mu_v)}{\partial\operatorname{\mu_v}(z_q')}-\varepsilon\cdot\frac{M}{M-1}\cdot\left(\frac{1}{(\pi_\varepsilon^*)_{oq}}-\frac{1}{(\pi_\varepsilon^*)_{op}}\right),
\end{equation*}
\end{small}

where $\pi_\varepsilon^*$ is the optimal primal solution to the entropy penalized OT problem, $z_j$ is any datapoint in $D_t$ other than $z_i$ or $z_k$, $z_o'$ is any datapoint in $D_v$ other than $z_p'$ or $z_q'$, $|D_t|=N$, and $|D_v|=M$.
\end{customthm}
\begin{proof}
Let $\mathcal{L}(\pi, f, g)$ and $\mathcal{L}_\varepsilon(\pi_\varepsilon, f_\varepsilon, g_\varepsilon)$ be the Lagrangian functions for original formulation and entropy penalized formulation between the datasets $D_t$ and $D_v$, respectively, which can be written as
\begin{equation*}
\begin{aligned}
    \mathcal{L}(\pi, f, g) = \langle\pi,c\rangle+\sum_{i=1}^N f_i\cdot(\pi_i'\cdot I_N-a_i)+\sum_{j=1}^M g_j&\cdot(I_M'\cdot\pi_j-b_j), \\
    \mathcal{L}_\varepsilon(\pi_\varepsilon, f_\varepsilon, g_\varepsilon) = \langle\pi_\varepsilon,c\rangle+\varepsilon\cdot\sum_{i=1}^{N}\sum_{j=1}^{M}\log\frac{(\pi_\varepsilon)_{ij}}{\mu_t(z_i)\cdot\mu_v(z_j)}&+\sum_{i=1}^N (f_\varepsilon)_i\cdot[(\pi_\varepsilon)_i'\cdot I_M-\mu_t(z_i))]\\
    &+\sum_{j=1}^M (g_\varepsilon)_j\cdot[I_N'\cdot(\pi_\varepsilon)_j-\mu_v(z_j)],
\end{aligned}
\end{equation*}
where $c^{N\times M}$ is the cost matrix consisting of distances between $N$ datapoints in $D_t$ and $M$ datapoints in $D_v$, $I_N=(1,1,...1)\in \mathbb{R}^{N \times 1}$ and  $I_M'=(1,1,...1)^T\in\mathbb{R}^{1\times M}$, $\pi$ and $(f,g)$ denote the primal and dual variables, and $\pi_i'$ and $\pi_j$ denote the $i^{th}$ row and $j^{th}$ column in matrix $\pi$, respectively.

The first-order necessary condition for optima in Lagrangian Multiplier Theorem 

gives that
\begin{equation*}
    \nabla \mathcal{L}_\pi(\pi^*, f^*, g^*)=0\quad \text{and}\quad  \nabla (\mathcal{L}_\varepsilon)_\pi((\pi_\varepsilon)^*, (f_\varepsilon)^*, (g_\varepsilon)^*)=0,
\end{equation*}
where $\pi^*$ and $(f^*, g^*)$ denote the optimal solutions to the primal and dual problems, respectively.
Thus, for any $i \in \{1,2,\ldots,N\}$ and $j \in \{1,2,\ldots,M\}$, we have
\begin{equation*}
\begin{aligned}
        \nabla \mathcal{L}_\pi(\pi^*, f^*, g^*)_{ij}&=c_{ij}+f^*_i+g^*_j=0,\\
        \nabla (\mathcal{L}_\varepsilon)_\pi(\pi_\varepsilon^*, f_\varepsilon^*, g_\varepsilon^*)_{ij}&=c_{ij}+\varepsilon\cdot\frac{1}{(\pi^*_\varepsilon)_{ij}}+(f_\varepsilon)^*_i+(g_\varepsilon)^*_j=0.
\end{aligned}
\end{equation*}

Subtracting, we have
\begin{equation*}
    \left[f_i^*-(f_\varepsilon)_i^*\right]+\left[g_j^*-(g_\varepsilon)_j^*\right]-\varepsilon\cdot\frac{1}{(\pi^*_\varepsilon)_{ij}}=0.
\end{equation*}

Then, for $\forall k\neq i \in \{1,2,...N\}$, we have
\begin{equation*}
    \left[f_k^*-(f_\varepsilon)_k^*\right]+\left[g_j^*-(g_\varepsilon)_j^*\right]-\varepsilon\cdot\frac{1}{(\pi^*_\varepsilon)_{kj}}=0.
\end{equation*}

Subtracting and reorganizing, we get
\begin{equation*}
    \left[(f_\varepsilon)_i^*-(f_\varepsilon)_k^*\right]=(f_i^*-f_k^*)-\varepsilon\cdot\left[\frac{1}{(\pi^*_\varepsilon)_{ij}}-\frac{1}{(\pi^*_\varepsilon)_{kj}}\right].
\end{equation*}

From the definition of the calibrated gradients in Eq.\ref{grd}, we have
\begin{equation*}
    \frac{\partial\operatorname{OT}(\mu_t,\mu_v)}{\partial\mu_t(z_i)}-\frac{\partial\operatorname{OT}(\mu_t,\mu_v)}{\partial\mu_t(z_k)}=\frac{N}{N-1}\left(f^*_i-f^*_k\right),
\end{equation*}
\begin{equation*}
    \frac{\partial\operatorname{OT_\varepsilon}(\mu_t,\mu_v)}{\partial\mu_t(z_i)}-\frac{\partial\operatorname{OT_\varepsilon}(\mu_t,\mu_v)}{\partial\mu_t(z_k)}=\frac{N}{N-1}\left[(f_\varepsilon)^*_i-(f_\varepsilon)^*_k\right].
\end{equation*}
Finally, subtracting and reorganizing, we have
\begin{small}
\begin{equation*}
\frac{\partial\operatorname{OT_\varepsilon}(\mu_t,\mu_v)}{\partial\mu_t(z_i)}-\frac{\partial\operatorname{OT_\varepsilon}(\mu_t,\mu_v)}{\partial\mu_t(z_k)}=\frac{\partial\operatorname{OT}(\mu_t,\mu_v)}{\partial\mu_t(z_i)}-\frac{\partial\operatorname{OT}(\mu_t,\mu_v)}{\partial\mu_t(z_k)}-\varepsilon\cdot\frac{N}{N-1}\cdot\left[\frac{1}{(\pi_\varepsilon^*)_{ij}}-\frac{1}{(\pi_\varepsilon^*)_{kj}}\right].
\end{equation*}
\end{small}
The proof for the second part of the Theorem is similar.
\begin{small}
\begin{equation*}
\frac{\partial\operatorname{OT_\varepsilon}(\mu_t,\mu_v)}{\partial\mu_v(z_p')}-\frac{\partial\operatorname{OT_\varepsilon}(\mu_t,\mu_v)}{\partial\mu_v(z_q')}=\frac{\partial\operatorname{OT}(\mu_t,\mu_v)}{\partial\mu_v(z_p')}-\frac{\partial\operatorname{OT}(\mu_t,\mu_v)}{\partial\mu_v(z_q')}-\varepsilon\cdot\frac{M}{M-1}\cdot\left[\frac{1}{(\pi_\varepsilon^*)_{op}}-\frac{1}{(\pi_\varepsilon^*)_{oq}}\right].
\end{equation*}
\end{small}
Then the proof is complete.
%\qedhere
\end{proof}

\newpage
\section{Additional Experimental Results}
\label{app:exp}

% on a range of  use cases. We cover datasets of different complexity.
% such as from a simple digit MNIST~\citep{deng2012mnist} dataset, through a medium traffic sign GTSRB~\citep{Stallkamp-IJCNN-2011} dataset and CIFAR100, ending with a higher complexity dataset, Tiny-ImageNet~\citep{deng2009imagenet}, which is a subset of ImageNet.

\begin{wrapfigure}{R
}{0.4\columnwidth}
\centering
% \vspace{-1.5em}
    \includegraphics[width=0.4\columnwidth]{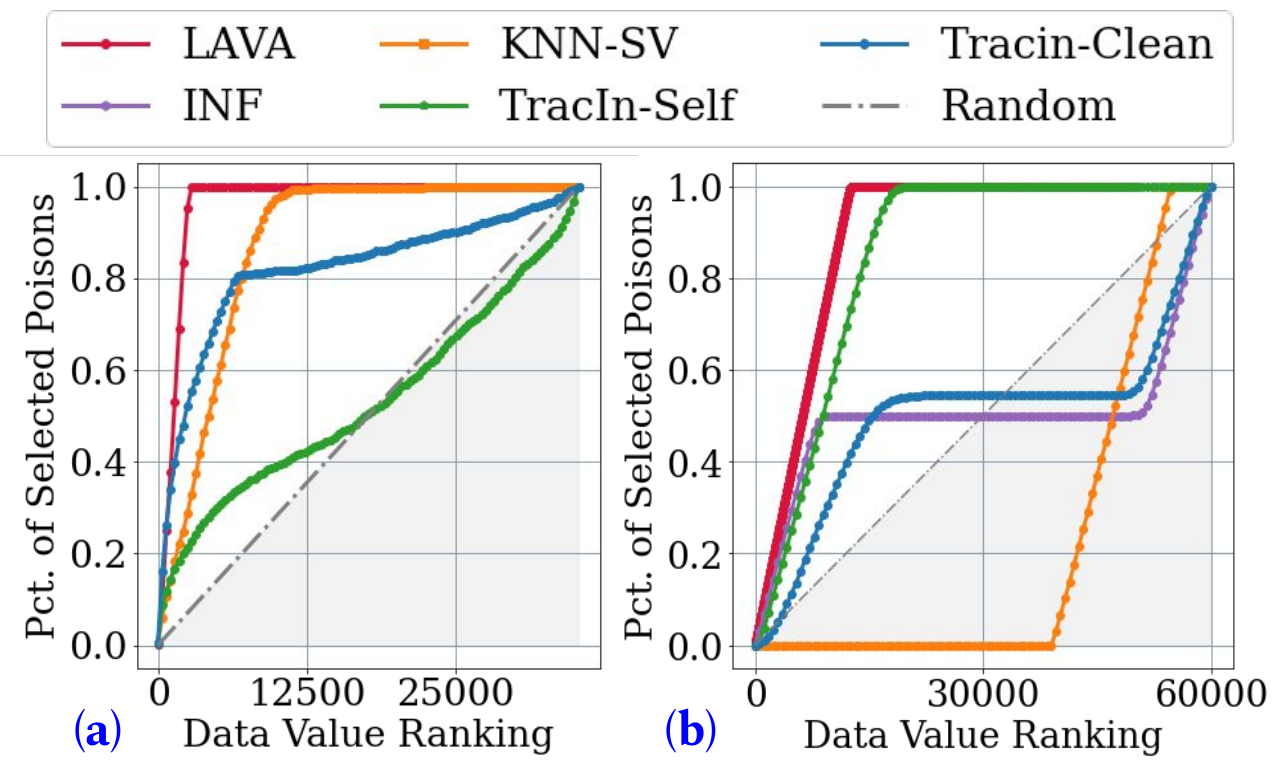}
    \vspace{-1.5em}
    \caption{(a) Blend backdoor attack detection on GTSRB dataset. Comparison with baselines. (b) Noisy feature detection on the digit MNIST dataset. Comparison with baselines.
    }
    \label{fig:back-gtsrb}
    \vspace{10em}
    % mm}
\end{wrapfigure}

% \begin{figure}[t!]
% \centering
%     \includegraphics[width=0.7\columnwidth]{figs/reprod-5.pdf}
%     \caption{(a) Blend backdoor attack on GTSRB dataset. Comparison with baselines. (b) Noisy feature detection on the digit MNIST dataset. Comparison with baselines.
%     }
%     \label{fig:back-gtsrb}
% \end{figure}

%IF possible do Tiny-ImageNet on one of the experiments at least.
%We have cifar100 - on irrelevant data

\subsection{Evaluating Data Valuation Use Cases on Diverse Datasets}
In the main text, we have focused our evaluation on CIFAR-10.
Here, we provide experiments to show effectiveness of $\AlgName$ on diverse datasets for detecting bad data.

\begin{wrapfigure}{R}{0.4\columnwidth}
    \vspace{-13em}
	\includegraphics[width=0.4\columnwidth]{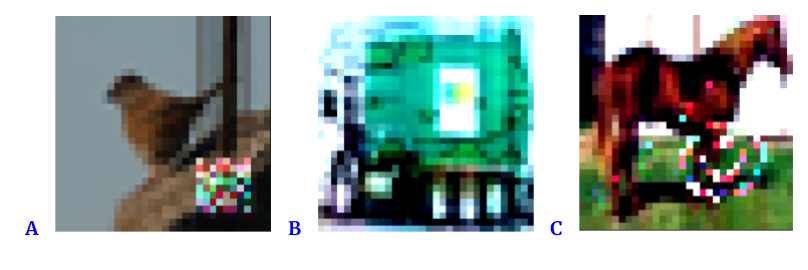}
	\vspace{-2em}
	\caption{Visualization of each backdoor attack: A) Trojan-SQ attack. B) Blend attack. C) Trojan-WM attack.
    }
    \vspace{-1em}
    \label{fig:back_patch}
    % \vspace{-1em}
\end{wrapfigure}

\textbf{Backdoor Attack Detection.} We evaluate another type of backdoor attack (Section~\ref{expe_back}), which is the Hello Kitty blending attack (Blend)~\citep{chen2017targeted} that mixes the target class sample with the Hello Kitty image, as illustrated in Figure~\ref{fig:back_patch} (B). We attack the German Traffic Sign dataset (GTSRB) on the target class $6$ by poisoning 1764 ($5\%$) samples of the whole dataset. Our method achieves the highest detection rate, as shown in Figure~\ref{fig:back-gtsrb}(a). In particular, the 5000 points with lowest data values contain all poisoned data based on the $\AlgName$ data values, while the second best method on this task, KNN-SV, can cover all poisoned examples with around 11,000 samples. Our algorithm performs especially well for this attack, since the label of poisoned data is changed to the target class and the patching trigger is large. Both the label and feature changes contribute to the increase of the OT distance and thus ease the detection.

\textbf{Noisy Feature Detection.}
% \begin{figure}[h!]
%     \centering
%     \includegraphics[width=0.5\columnwidth]{figs/noisy-mnist.png}
%     \caption{Noisy feature detection on the digit MNIST dataset. Comparison with baselines.
%     }
%     \label{fig:noisy-mnist}
% \end{figure}
Here, we show the usage of $\AlgName$ on the MNIST dataset where $25\%$ of the whole dataset is contaminated by feature noise. 
% Although, one of the baselines, \textsl{TraceIn-Self}, performs well on this experiment as shown in Figure~\ref{fig:back-gtsrb}(b), o
Our method still outperforms all the baselines by detecting all noisy data within first 14,000 samples, which is 5,000 less than the best baseline would require, which is shown in Figure~\ref{fig:back-gtsrb}(b). 
% Since the feature noise is strong even without the label change, our method detects well the noises.
\begin{wrapfigure}{R
}{0.41\columnwidth}
\vspace{-10.5mm}
\centering%
    \vspace{-11em}
	\includegraphics[width=0.35\columnwidth]{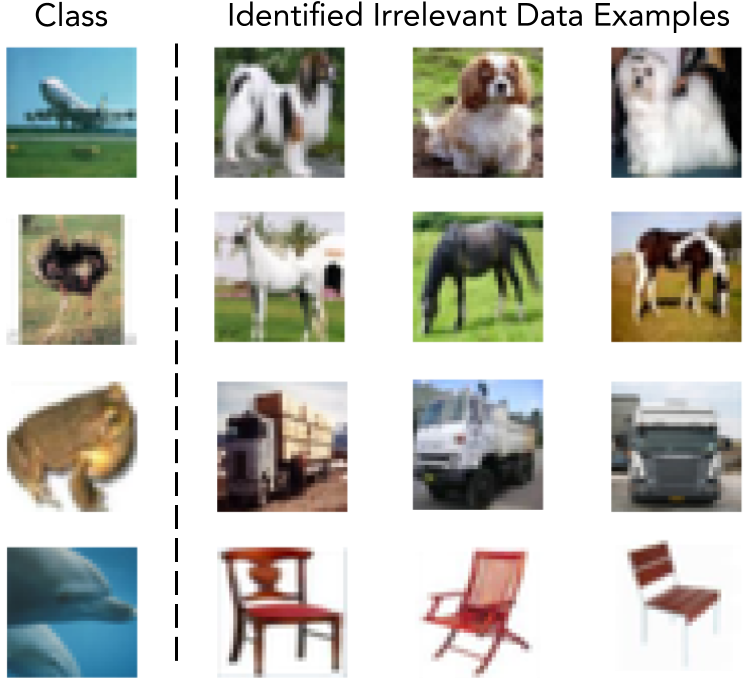}
	\caption{ Visualization of irrelevant data detection within the CIFAR100 dataset. The \textbf{left} column is one example of the target class and the images on the \textbf{right} columns are selected irrelevant data in the corresponding classes detected by $\AlgName$.
    }
    \label{fig:irrel}
    \vspace{5em}
\end{wrapfigure}

% \textbf{Mislabeled Data Detection.} \ruoxi{need to revise after the figure is uploaded.}
% % In this experiment, we want to show the scalability of our method on large datasets. Specifically, w
% We perform the mislabeling detection on the STL10 dataset, where we shuffle the labels of $20\%$ of all 5,000 samples in the training set. In Figure ~\ref{fig:back_patch} (C), we can observe that our method performs comparably to other baselines in the given scenario.

% We perform the mislabeling detection on the ImageNet-100 dataset, where we shuffle the labels of $25\%$ of all 100,000 samples in the training set.
% % $\AlgName$ finishes all computation in around 1 GPU hour ($+$30 minutes, depending on the system). 
% In Figure~\ref{fig:back-gtsrb}(c), we can observe the performance of our method compared with three of the baselines. Other baselines are not included in the figure, since they can not finish the task within 5 hours.
% % \textcolor{red}{within the limited time}. 
% Our method enjoys the nearly linear time complexity and memory overhead, which allows to execute large scale datasets.

\textbf{Irrelevant Data.} We perform another irrelevant data detection experiment and focus on the CIFAR100 dataset. In Figure~\ref{fig:irrel}, we illustrate some of the irrelevant samples detected by $\AlgName$. Intuitively, irrelevant data in the class should be easily detected by $\AlgName$, since the images are far from the representative of the class and increasing the probability mass associated with these images leads to larger distributional distance to the clean validation data.
% if have enough time, I will put more results on cifar10

% \vspace{-4em}
%     \includegraphics[width=0.4\columnwidth]{figs/mislabel-fe.png}
%     \caption{Comparison between different feature embedder architectures on inspecting 50k samples from CIFAR-10 with $25\%$ mislabeled data.
%     }
%     \vspace{-2em}
%     \label{fig:mislabel-fe}
% \begin{figure}[H]
%     \centering%
% 	\includegraphics[width=0.7\linewidth]{figs/irrelevant.png}
% 	\caption{ Visualization of irrelevant data detection within the CIFAR100 dataset. The \textbf{left} column is one example of the target class and the images on the \textbf{right} columns are selected irrelevant data in the corresponding classes detected by $\AlgName$.
%     }
%     \label{fig:irrel}
% \end{figure}

\subsection{Ablation Study}

% We have specified the hyperparameters we used for experiments in the main paper. Here, as promised, we go through 

We perform an ablation study on validation size and on the hyperparameters in our method, where we provide insights on the impact of setting changes. We use the mislabeled detection use case and the CIFAR-10 dataset as an example setting for the ablation study.

% the effect of specific changes on the mislabeling settings from Section~\ref{expe} in the main paper on the CIFAR10 dataset.

% Do both: Mislabeling and Poisoning Experiments

\subsubsection{Validation Size}
\label{app:val_size}

% \begin{wrapfigure}{R}{0.3\columnwidth}
%     \vspace{-1em}
%     \includegraphics[width=0.3\columnwidth]{figs/mislabel500.png}
%     \caption{Detection rate by various methods on CIFAR-10 50k samples with $25\%$ mislabeled data using validation size of 500.
%     }
%     \label{fig:valid-size-500}
% \end{wrapfigure}

% \vspace{2em}

\begin{figure}[t!]
    \vspace{1em}
    \centering
    \includegraphics[width=0.85\columnwidth]{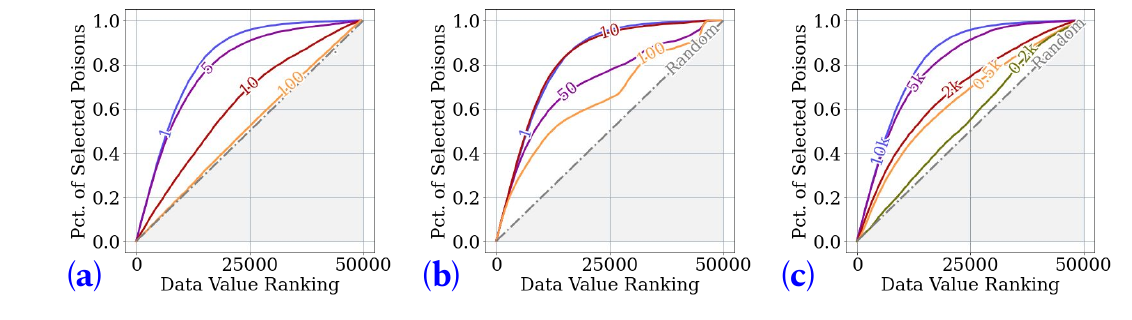}
    % \subfigure[] {\includegraphics[width=0.25\columnwidth]{figs/feat-weights.png}}
    % \subfigure[]{\includegraphics[width=0.25\columnwidth]{figs/label-weight-2.png}}
    % \subfigure[]{\includegraphics[width=0.25\columnwidth]{figs/valid-size.png}}
    \caption{(a) Comparison between different feature weights on the performance of mislabeled data in CIFAR-10. (b) Comparison between different label weights on the performance of mislabeled data in CIFAR-10. (c) Comparison between different validation sizes on inspecting 50k samples from CIFAR-10 with $25\%$ mislabeled data.
    }
    \vspace{-1em}
    \label{fig:misl-lbl-wgt}
\end{figure}

\begin{wrapfigure}{R}{0.3\columnwidth}
% \vspace{-5em}
% \centering
    % \vspace{-1em}
    \includegraphics[width=0.3\columnwidth]{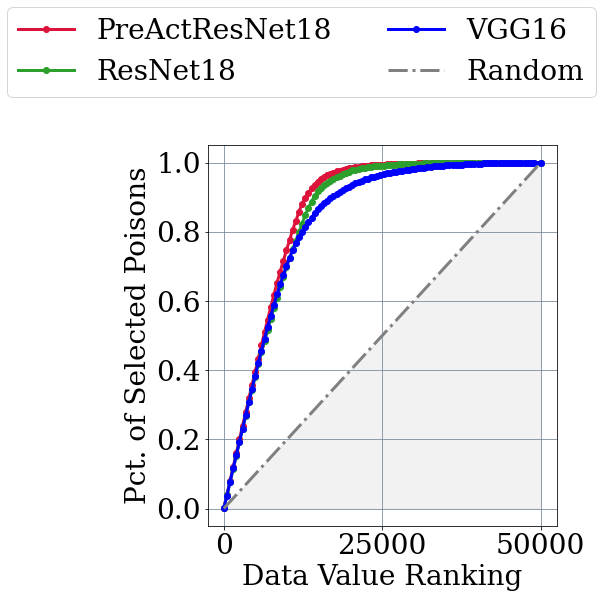}
    \vspace{-1em}
    \caption{Comparison between different feature embedder architectures on inspecting 50k samples from CIFAR-10 with $25\%$ mislabeled data.
    }
    \vspace{-1em}
    \label{fig:mislabel-fe}
    \vspace{2em}
    \includegraphics[width=0.3\columnwidth]{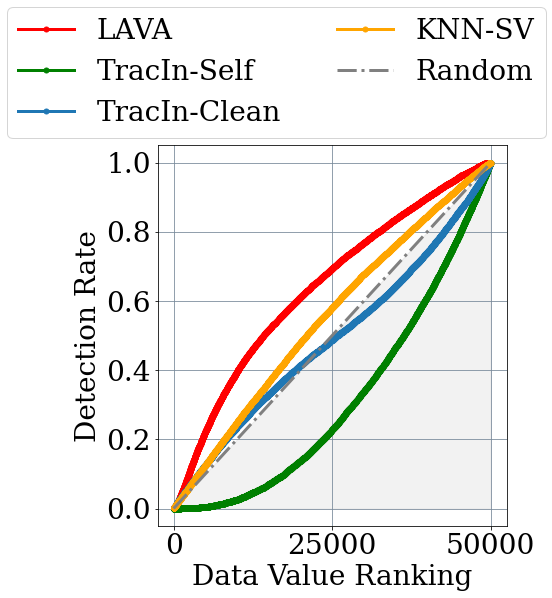}
    \vspace{-1.2em}
    \caption{Detection rate by various methods on mislabeled CIFAR-10 using validation size of 500.
    }
     \vspace{0em}
    \label{fig:valid-size-500}
\end{wrapfigure}

For all the experiments in the main text, we use the validation set of size 10,000. Naturally, we want to examine the effect of the size of the validation set on the detection rate of mislabeled data. In Figure~\ref{fig:misl-lbl-wgt} (c), we illustrate the performance on the detection rate with smaller validation data sizes: $200$, $500$, $2,000$, and $5,000$. We observe that even reducing the validation set by half to $5,000$ can largely maintain the detection rate performance. Small validation sets ($200$, $500$, $2,000$) degrade the detection rate by more than $50\%$. Despite the performance degradation, our detection performance with these small validation sizes is in fact comparable with the baselines in Figure~\ref{fig:all_detection} IV.(a) that leverage the full validation size of $10,000$. Additionally, when restricting $\AlgName$ and other baselines to validation set of 500 samples, our method is better than the best baseline for detecting mislabeled data in the 50k CIFAR-10 samples with $25\%$ being mislabeled as shown in Figure~\ref{fig:valid-size-500}.

% \vspace{1em}

% \begin{figure}[t!]
%     \vspace{1em}
%     \centering
%     \includegraphics[width=0.9\columnwidth]{iclr2023/figs/valid_model.pdf}
%     % \subfigure[] {\includegraphics[width=0.25\columnwidth]{figs/feat-weights.png}}
%     % \subfigure[]{\includegraphics[width=0.25\columnwidth]{figs/label-weight-2.png}}
%     % \subfigure[]{\includegraphics[width=0.25\columnwidth]{figs/valid-size.png}}
%     \caption{Detection performance comparison between LAVA and the model trained on validation data of size $10,000$ on various use cases in CIFAR-10.\kang{which paragraph is it corresponding to? \edit{Paragraph C2.1}}
%     }
%     % \vspace{-5em}
%     \label{fig:validation_model}
% \end{figure}

\subsubsection{Feature Weight}

Recall the class-wise Wasserstein distance is defined with respect to the following distance metric: \add{$\C(  (x_t, y_t), (x_v, y_v) ) = \metric(\xt, \xv) + c W_{\metric}(  \mut(\cdot | y_t) , \muv(\cdot | y_v) )$}. Actually, one can change the relative weight between feature distance \add{$\metric(\xt, \xv)$} and the label distance \add{$W_{\metric}( \mut(\cdot | y_t), \muv(\cdot | y_v) )$}. Here,
% In the main paper, we have focused on uniform weights for feature and label distances. We 
we show the effect of upweighting the feature distance, while keeping the label weight at 1 and the results are illustrated in Figure~\ref{fig:misl-lbl-wgt} (a). As we are moving away from uniform weight, the performance on detection rate is decreasing with larger feature weights. With feature weight of $100$, our method performs similarly as the random detector. Indeed, as we increase weight on the features, the weight on the label distance is decrease\add{d}. As the weight reaches 100, our method performs similarly as the feature embedder without knowing label information and hence, the mislabeled detection performance is comparable to the random baseline.

\vspace{1em}
\subsubsection{Label Weight}

Next, we shift focus to label weight. We examine the effect of upweighting the label distance, while keeping the feature weight at 1. In Figure~\ref{fig:misl-lbl-wgt} (b), as the label weight increases, the detection rate performance deteriorates. When we increase the label distance, the feature information becomes neglected, which is not as effective as the balanced weights between feature and label distances.

% \subsubsection{Ablation Study: Entropy Regularizer vs Performance and Time}
% No effect- do we want to mention? Or exclude? Not explained in the main papaer

% \vspace{1em}

\subsubsection{Feature Embedder}
\label{app:feat_emb}
% \ruoxi{Use one application, e.g., mislabeled data, to perform all ablation study. if needed, we can add more applications to specific ablation study}

\begin{wrapfigure}{R}{0.3\columnwidth}
    \vspace{-14.0em}
    \includegraphics[width=0.3\columnwidth]{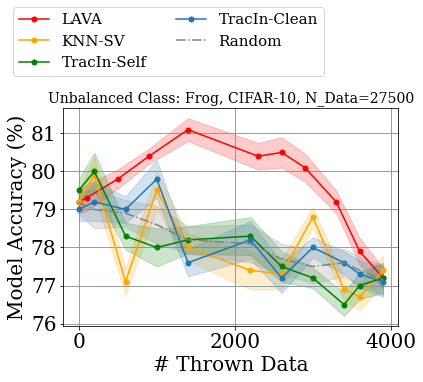}
    \vspace{-1.5em}
    \caption{Comparison of various methods on rebalancing an unbalanced dataset on CIFAR-10 with the class \emph{Frog} being unbalanced. 
    }
    \vspace{1em}
    \label{fig:unbalance}
    
    % \begin{wrapfigure}{R}{0.3\columnwidth}
    
    \includegraphics[width=0.3\columnwidth]{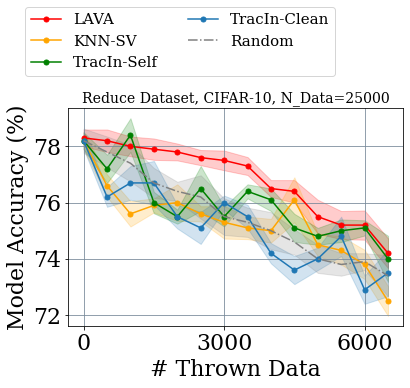}
    \vspace{-2em}
    \caption{Comparison of various methods on reducing a dataset size based on valuation of datapoints on CIFAR-10.
    }
    \vspace{-3em}
    \label{fig:reduce}
    
% \end{wrapfigure}
% \vspace{0.5em}
% \caption{Wasserstein distance behavior under dataset direct duplication and its near duplicates.}%\yi{re-scale your table to mach the text size of Figure 2}}
% \label{table:time-scale}
\end{wrapfigure}

\begin{figure}[t!]
    \vspace{1em}
    \centering
    \includegraphics[width=0.9\columnwidth]{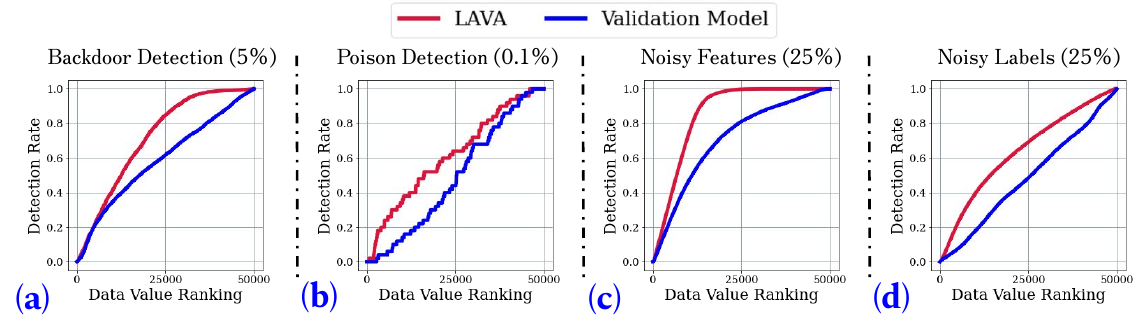}
    % \subfigure[] {\includegraphics[width=0.25\columnwidth]{figs/feat-weights.png}}
    % \subfigure[]{\includegraphics[width=0.25\columnwidth]{figs/label-weight-2.png}}
    % \subfigure[]{\includegraphics[width=0.25\columnwidth]{figs/valid-size.png}}
    \caption{\edit{Detection performance comparison between LAVA and the model trained on validation data of size $500$ on various use cases in CIFAR-10.}
    }
    % \vspace{-5em}
    \label{fig:validation_model500}
    \vspace{1em}
    \centering
    \includegraphics[width=0.9\columnwidth]{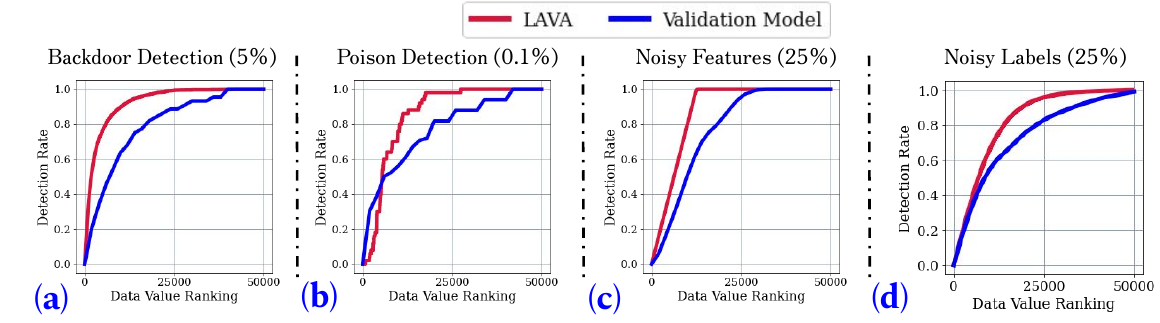}
    % \subfigure[] {\includegraphics[width=0.25\columnwidth]{figs/feat-weights.png}}
    % \subfigure[]{\includegraphics[width=0.25\columnwidth]{figs/label-weight-2.png}}
    % \subfigure[]{\includegraphics[width=0.25\columnwidth]{figs/valid-size.png}}
    \caption{\edit{Detection performance comparison between LAVA and the model trained on validation data of size $10,000$ on various use cases in CIFAR-10.}
    }
    % \vspace{-5em}
    \label{fig:validation_model}
\end{figure}
% \begin{wrapfigure}{R}{0.4\columnwidth}
%     \includegraphics[width=0.4\columnwidth]{figs/feat-weights.png}
%     \caption{Comparison between different feature weights on the performance of mislabeled data in CIFAR-10.
%     }
%     \label{fig:misl-ft-wgt}
% \end{wrapfigure}

We use feature embedder to extract features for the feature distance part in our method. 
% Since our validation set consists of samples representative of qualities that the buyer desires, we
We train the feature embedder on the accessible validation set until the convergence on the train accuracy. Different architectures of the embedder might be sensitive to different aspects of the input and thus result in different feature output. Nevertheless, as we observe in Figure~\ref{fig:mislabel-fe}, the detection performance associated with different model architectures of feature embedder is similar. Hence, in practice, one can flexibly choose the feature embedder to be used in tandem with our method as long as it has large enough capacity. \edit{Furthermore, we note that that these feature embedders have not learned the clean distribution from the validation data, e.g. in CIFAR-10 the model trained on $10K$ validation data achieves only around $65\%$ accuracy on $50K$ clean datapoints and the model trained on $500$ validation data achieves around $25\%$ accuraracy. We additionally show in Figure~\ref{fig:validation_model500},\ref{fig:validation_model} that our method significantly out-\\
performs the PreActResNet18 model trained directly on validation \\ 
data of size $500$ and $10K$ in
detecting bad data, which clearly dis-\\
tinguishes $\AlgName$ from simple feature embedders.}
% as long as itThus, we want to convey a message that our method is flexible of the feature embedder model architecture as long as it is ``well-trained'' on the validation data.

\subsection{\add{Balancing Unbalanced Dataset}}

Although machine leaning practitioners might be using clean data for training a model, the dataset can be often unbalanced which can lead to model performance degradation~\citep{thai2009improving}. To recover higher model accuracy, we can rebalance unbalanced datasets by removing points that cause such disproportion. We showcase how $\AlgName$ effectively rebalance the dataset by removing points with poor values and keeping points with best values. We consider a CIFAR-10 dataset with a class \emph{Frog} being unbalanced and containing $5,000$ samples while other classes have only half as much (i.e. $2,500$ samples). In Figure~\ref{fig:unbalance}, we demonstrate the effectiveness of $\AlgName$ valuation which not only reduces the dataset by removing poor value points but also improves the model accuracy. While at the same time other valuation methods were not able to steadily increase the model accuracy and quickly downgraded the model performance, which in turn shows an even stronger effectiveness of our method.

    % \includegraphics[width=0.3\columnwidth]{figs/reduce_nip.png}
    % \vspace{-1em}
    % \caption{Comparison of various methods on reducing a dataset size based on valuation of datapoints on CIFAR-10.
    % }
    % \vspace{-1em}
    % \label{fig:reduce}

\begin{wrapfigure}{R}{0.3\columnwidth}
    \vspace{-3em}
    \includegraphics[width=0.3\columnwidth]{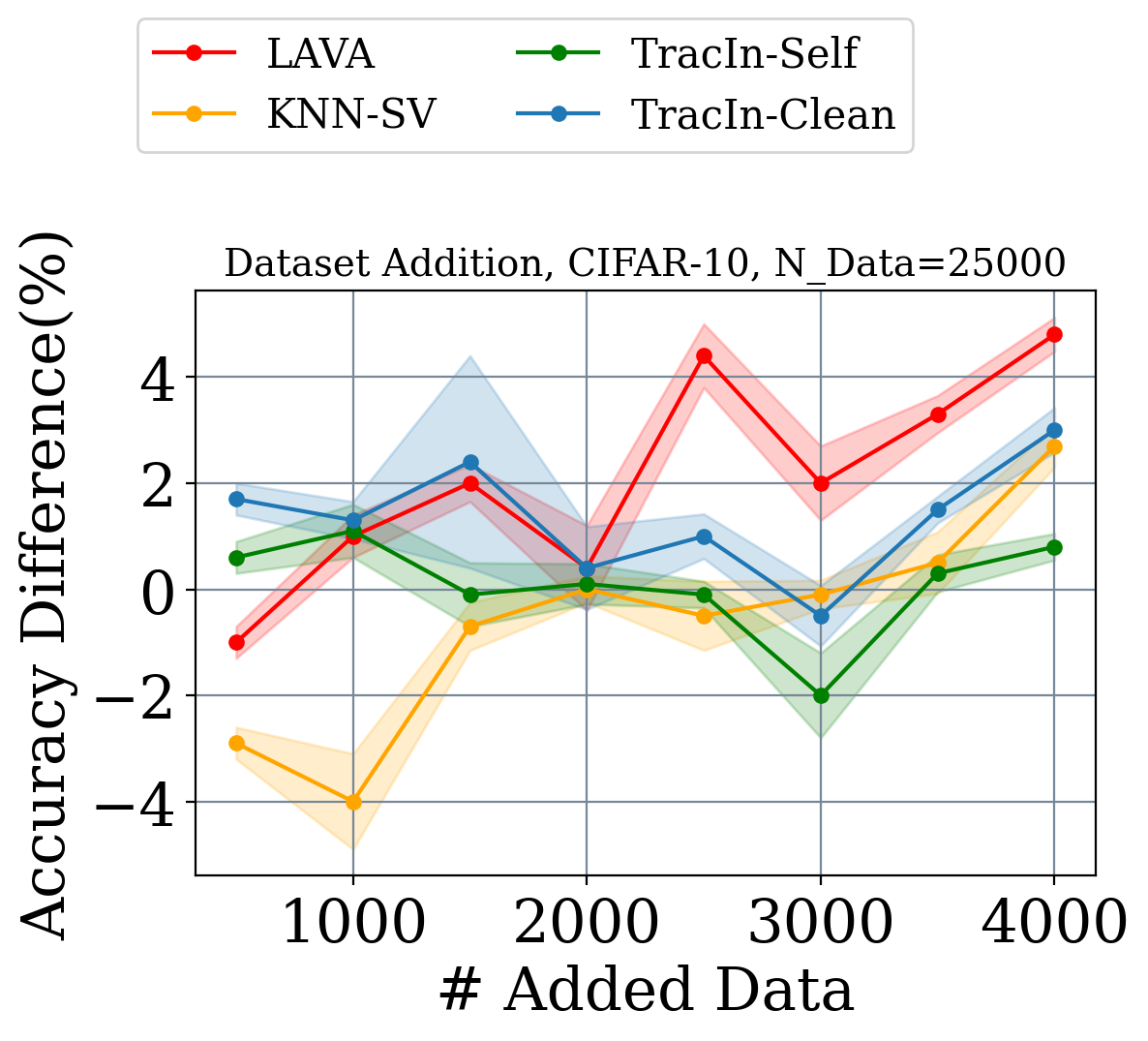}
    \vspace{-1.7em}
    \caption{Comparison of various methods on data summarization based on valuation of datapoints on CIFAR-10.
    }
    \vspace{0.5em}
    \label{fig:addition}

    \includegraphics[width=0.3\columnwidth]{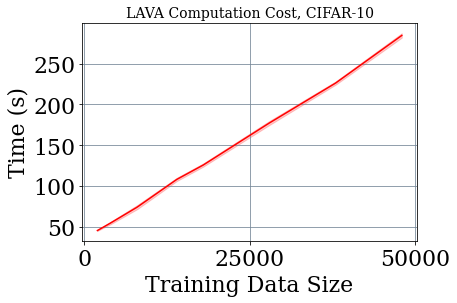}
    \vspace{-2.5em}
    \caption{Near-linear time complexity of $\AlgName$ shown on CIFAR-10. 
    }
    % \vspace{-5em}
    \label{fig:nearlinear}
 \vspace{-3em}
 
\end{wrapfigure}
\subsection{Reducing Training Set Size}
\label{subsection:reduce_set}

With the growing size of the training dataset, the computation cost and memory overhead naturally increase which might deem impossible for some practitioners with limited resources to train a model. Therefore, the ability to reduce the training dataset size~\citep{sener2018active} will free up some computation burden and thus allow ones with limited resources to fully appreciate the model training process. Motivated by the given challenge, we want to leverage our data valuation method to significantly decrease the training dataset size while maintaining the model performance. Similarly as in the previous section, the idea is to keep a subset of datapoints with best values and remove poor valued ones. To demonstrate the effectiveness of our $\AlgName$'s valuation, we perform such a task on a clean CIFAR-10 dataset with $2,500$ samples from each class and compare with other data valuation methods. As presented in Figure~\ref{fig:reduce}, it is demonstrated that the performance is well maintained even with smaller subsets of the original dataset. Remarkably, even reducing a clean training set (25,000 samples) by $15\%$ based on our method’s valuation, the performance can still stay relatively high while outperforming other valuation baselines.

\subsection{Data Summarization}

With growing dataset sizes, grows the space needed to store data. Thus, the buyer often would like to decrease the dataset to minimize resources but to retain the performance. Unlike reducing training set size as provided in Section~\ref{subsection:reduce_set}, in this experiment, we will select a smaller, representative subset of the whole dataset that can maintain good performance. To measure the performance of each subset, we measure the validation performance of the model trained on that subset subtracted by the validation performance of the model trained on a random subset of the same size, the experiment which is performed in~\cite{kwon2021beta}. In Figure~\ref{fig:addition}we can observe that our method can select a small subset that performs better than the subsets chosen by the baseline methods most of the time. 

\subsection{Scalability Experiment}

In the main paper, we have demonstrated time complexity comparison between $\AlgName$ and other valuation methods. We have reported runtime comparisons only for 2,000 test samples as this is the scale existing methods can solve in a not excessively long time (within a day). It showcases the advantageous computing efficiency that the proposed approach enjoys over other methods. We further want to emphasize the computational efficiency of $\AlgName$ and demonstrate computation efficiency on a larger scale dataset (100,000 samples) with higher dimensions, ImageNet-100. Additionally, we evaluate other baselines which are able to finish within a day of computation to highlight the advantage of our method as presented in Table~\ref{table:time-scale}. Moreover, we highlight the near-linear time complexity of $\AlgName$ on CIFAR-10, which shows practical computation efficiency of our method as shown in Figure~\ref{fig:nearlinear}.

% to show the general tendency of the Wasserstein distance being the upper bound of the model performance, which agrees with the theoretical results in our paper.

\subsection{Generalization to Other Types of Backdoor Attacks}

% \begin{wraptable}{R}{0.30\columnwidth}

% % \vspace{-1em}
% \begin{tabular}{l|r}
% \multicolumn{1}{c}{\textbf{Dataset}}                      & \multicolumn{1}{c}{\textbf{ Time}}  \\
% \toprule
% \AlgName                           & 1 hr 54 min                                   \\
% KNN-SV                              & 4 hr 21 min                                    \\
% TracIn-Clean                                & 7 hr 50 min                                    \\
% TracIn-Self         & 7 hr 51 min                                       
% \end{tabular}
% \vspace{0.5em}
% \caption{Wasserstein distance behavior under dataset direct duplication and its near duplicates.}%\yi{re-scale your table to mach the text size of Figure 2}}
% \label{table:time-scale}
% \end{wraptable}

% \begin{figure}
%     \centering
% 	\includegraphics[width=0.99\linewidth]{figs/backdoors.png}
% 	\caption{Visualization of each backdoor attack: A) Trojan-SQ attack. B) Blend attack. C) Trojan-WM attack.
%     }
%     \label{fig:back_patch}
% \end{figure}

\begin{wrapfigure}{r}{0.3\columnwidth}
\centering
% \vspace{-13.5em}
% \vspace{1em}

% \begin{tabular}{l|r}
% \multicolumn{1}{c}{\textbf{Method}}                      & \multicolumn{1}{c}{\textbf{ Time}}  \\
% \toprule
% \AlgName                           & 1 hr 54 min                                   \\
% KNN-SV                              & 4 hr 21 min                                    \\
% TracIn-Clean                                & 7 hr 50 min                                    \\
% TracIn-Self         & 7 hr 51 min                                       
% \end{tabular}
% \vspace{0.5em}
% \caption{Comparison of runtime between various methods needed to valuate ImageNet-100.}%\yi{re-scale your table to mach the text size of Figure 2}}
% \label{table:time-scale}

% \vspace{1em}

%     \begin{tabular}{l|r}
% \multicolumn{1}{c}{\textbf{Dataset}}                      & \multicolumn{1}{c}{\textbf{ Time}}  \\
% \toprule
% \AlgName                           & 1 hr 54 min                                   \\
% KNN-SV                              & 4 hr 21 min                                    \\
% TracIn-Clean                                & 7 hr 50 min                                    \\
% TracIn-Self         & 7 hr 51 min                                       
% \end{tabular}
\vspace{-2em}

    \includegraphics[width=0.28\columnwidth]{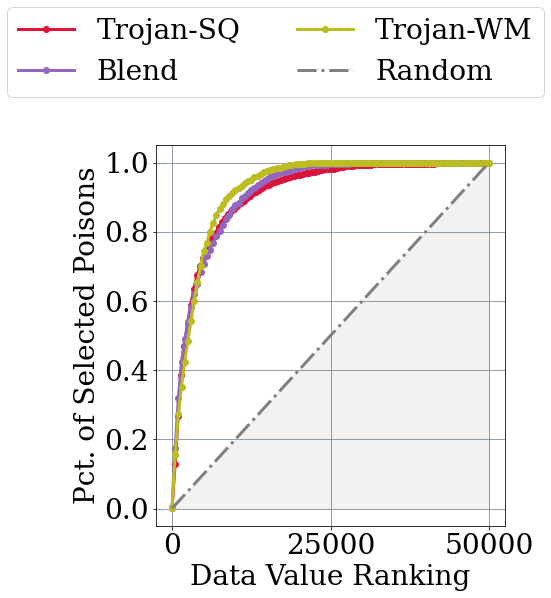}
     \vspace{-1em}
    \caption{Detection rate of various backdoor attacks by $\AlgName$.
    }
    % \vspace{8em}
    \label{fig:back_grad}

\end{wrapfigure}

As we have provided the results of the Trojan square attack (Trojan-SQ)~\citep{liu2017trojaning} in Section~\ref{expe}, we now apply $\AlgName$ to other backdoor attacks, which are Hello Kitty blending attack (Blend)~\citep{chen2017targeted} and Trojan watermark attack (Trojan-WM)~\citep{liu2017trojaning}, and evaluate the efficacy of our method in detecting different types of backdoor attacks. 
We simulate these attacks by selecting the target class \textsl{Airplane} and poisoning $2,500$ ($5\%$) samples of the CIFAR-10 dataset of size $50,000$. The backdoor trigger adopted in each attack is portrayed in Figure~\ref{fig:back_patch}. In Figure~\ref{fig:back_grad}, we observe that our method can achieve superior detection performance on all the attacks considered. The reason is that despite the difference in trigger pattern, all of these attacks modify both the label and the feature of a poisoned image and thus result in the deviation of our distributional distance that is defined over the product space of feature and label.
% similar performance of backdoor detection even between attacks. 
% The reason is that their label change is the same which is reflected in the label distance of our method. The difference between attacks is the patch that is applied to the images. 

\begin{wraptable}{r}{0.35\columnwidth}

\vspace{-2em}
\begin{tabular}{l|r}
\multicolumn{1}{c}{\textbf{Method}}                      & \multicolumn{1}{c}{\textbf{ Time}}  \\
\toprule
\AlgName                           & 1 hr 54 min                                   \\
KNN-SV                              & 4 hr 21 min                                    \\
TracIn-Clean                                & 7 hr 50 min                                    \\
TracIn-Self         & 7 hr 51 min                                       
\end{tabular}
\vspace{0.5em}
\caption{Comparison of runtime between various methods needed to valuate ImageNet-100.}%\yi{re-scale your table to mach the text size of Figure 2}}
\label{table:time-scale}

    \vspace{0em}
% \begin{tabular}{r|rr}
% \multicolumn{1}{c}{\textbf{Dataset}}                      & \multicolumn{2}{c}{\textbf{ OT Dist}}  \\
% \multicolumn{1}{c}{\textbf{Size}} & Direct  & Near                      \\ 
% \toprule
% $5000          $                             & $195.64$ & $+0.98$                                     \\
% $2 \times 5000    $                               & $+0.00$ & $+0.98$                                     \\
% $3 \times 5000 $                                  & $+0.00$ & $+0.98$                                     \\
% $4 \times 5000  $                                 & $+0.00$ & $+0.98$                                     \\
% $4.5 \times 5000  $                               & $+0.07$ & $+0.98$                                     \\
% $5 \times 5000    $                               & $+0.00$ & $+0.98$                                    
% \end{tabular}
% % \vspace{1em}
% \caption{Class-wise Wasserstein distance behavior under dataset direct duplication and its near duplicates.}%\yi{re-scale your table to mach the text size of Figure 2}}
% \label{table:dup}

\end{wraptable}

\subsection{Implications of the Proposed Data Valuation Method to Real-World Data Marketplaces}
% \begin{wraptable}{R}{0.35\columnwidth}

% \begin{tabular}{r|rr}
% \multicolumn{1}{c}{\textbf{Dataset}}                      & \multicolumn{2}{c}{\textbf{ Wasserstein Dist}}  \\
% \multicolumn{1}{c}{\textbf{Size}} & Direct  & Near                      \\ 
% \toprule
% $5000          $                             & $195.64$ & $+0.98$                                     \\
% $2 \times 5000    $                               & $+0.00$ & $+0.98$                                     \\
% $3 \times 5000 $                                  & $+0.00$ & $+0.98$                                     \\
% $4 \times 5000  $                                 & $+0.00$ & $+0.98$                                     \\
% $4.5 \times 5000  $                               & $+0.07$ & $+0.98$                                     \\
% $5 \times 5000    $                               & $+0.00$ & $+0.98$                                    
% \end{tabular}
% \vspace{1em}
% \caption{Wasserstein distance behavior under dataset direct duplication and its near duplicates.}%\yi{re-scale your table to mach the text size of Figure 2}}
% \label{table:dup}
% \end{wraptable}

One concern in the real-world data marketplace is that data is freely replicable. However, replicates of data introduce no new information and therefore the prior work has argued that a data utility function should be robust to direct data copying~\citep{xu2021validation}.
% A limitation of using validation performance as the utility function is that it could be sensitive to data replication. For instance, when validation data is class-imbalanced, replicating the points in the dominating class can potentially increase the model performance by biasing the model prediction to that class. 
One advantage of using the class-wise Wasserstein distance to measure data utility is that it is \emph{robust to duplication}. 
% for a dataset metric to be useful in the market is the ability to ignore or remove duplicate data and its near duplicates in a dataset . 
Our method by its natural distributional formulation will ignore duplicates sets. As shown in Table~\ref{table:dup}, although we have repeated the set even five times more than the original source set, the distance remains the same. Additionally, with small noise changes in the features, the distance metric is barely affected. Another concern in the real-world marketplace is that one might find a single data that has highest contribution and duplicate it to maximize the profit. However, again due to the nature of our distributional formulation, duplicating a single point multiple times would increase the distance between the training and the validation set due to the imbalance in training distribution caused by copying that point.

\subsection{Detailed Experimental Settings}
% \kang{format here}

\textbf{Datasets and Models.} Table~\ref{table:data_model} summarizes the details of the dataset, the models, as well as their licenses adopted in our experiments. 
% For $\AlgName$ implementation, we use the current assets with public licenses:

% \paragraph{Datasets.}
% \begin{itemize}
%     \item CIFAR-10/100 ~\citep{krizhevsky2009learning} - licensed under the MIT license. Downloaded from: https://www.cs.toronto.edu/~kriz/cifar.html.
%     \item STL10 ~\citep{coates2011analysis} - licensed under public, non-specified license. Downloaded from: https://cs.stanford.edu/~acoates/stl10/.
%     \item Tiny-ImageNet ~\citep{deng2009imagenet} - licensed under public, non-specified license. Downloaded from: https://image-net.org/.
% \end{itemize}
% \paragraph{Models.}
% \begin{itemize}
%     \item PreActResNet18/34 ~\citep{he2016identity} - licensed under MIT license.
%     \item ResNet18/50 ~\citep{he2016deep} - licensed under MIT license.
%     \item GoogLeNet ~\citep{szegedy2015going} - licensed under MIT license.
% \end{itemize}
% \paragraph{Packages.}
% \begin{itemize}
%     \item pytorch ~\citep{NEURIPS2019_9015} - licensed under BSD license.
%     \item otdd ~\citep{alvarez2020geometric} - licensed under MIT license.
%     \item geomloss ~\citep{feydy2019interpolating} - licensed under MIT license.
%     \item numpy ~\citep{harris2020array} - licensed under liberal BSD license.
% \end{itemize}
\begin{table}
\centering
\caption{Summary of datasets and models and their licenses used for experimental evaluation. (Note: ``Detail'' column implicitly refers to training data, unless explicitly noted.)} %\ruoxi{double check typos and formatting issues}}
\label{table:data_model}
\vspace{1em}
\begin{adjustbox}{width=\textwidth,totalheight=\textheight,keepaspectratio}
\begin{tabular}{lllllll}
\toprule
\begin{tabular}[c]{@{}l@{}}
\textbf{Experiment }\\\textbf{Type}\end{tabular} & \textbf{Dataset}                                       & \textbf{License}   & \textbf{Model}                                            & \begin{tabular}[c]{@{}l@{}}\textbf{License}\\\end{tabular} & \begin{tabular}[c]{@{}l@{}}\textbf{Train/}\\\textbf{Valid Size}\end{tabular}        & \textbf{Detail}                                                                         \\ 

\midrule
Wasserstein vs  & STL10    & \cellcolor[HTML]{C0C0C0}{\color[HTML]{FFFFC7} Not Applicable} & GoogLeNet  & MIT   & 5K/8K   &  $0\%,2\%,5\%,10\%,15\%$   Mislabeled\\
\begin{tabular}[c]{@{}l@{}}Model \\
Performance (Fig~\ref{fig:theory})\end{tabular}                 & CIFAR10      & MIT     & ResNet18    & MIT   & 50K/10K  &  $0\%,2\%,5\%,10\%,15\%$      Mislabeled                                                                    \\
  & CIFAR100   & MIT     & ResNet50   & MIT  & 50K/10K &  $0\%,2\%,5\%,10\%,15\%$ Mislabeled                                                  \\
  
  \midrule
\begin{tabular}[c]{@{}l@{}}Backdoor: \\Trojan Sq (Fig~\ref{fig:all_detection})\end{tabular}               & CIFAR10     & MIT   & ResNet18  & MIT   & 50K/10K   & 5\% Poisoned \\

\midrule
\begin{tabular}[c]{@{}l@{}}Poisoning: \\Collision (Fig~\ref{fig:all_detection})\end{tabular}        & CIFAR10  & MIT   & ResNet18   & MIT  & 50K/10K  & 0.1\% Poisoned   \\

\midrule
Noisy Feature (Fig~\ref{fig:all_detection}) & CIFAR10   & MIT & ResNet18  & MIT & 50K/10K   & 25\% Noisy Data \\

\midrule
Noisy Labels  (Fig~\ref{fig:all_detection}) & CIFAR10   & MIT   & ResNet18  & MIT & 50K/10K  & 25\% Noisy Data\\

\midrule
Irrelevant Data  (Fig~\ref{fig:data_irl})& CIFAR10 & MIT  & \begin{tabular}[c]{@{}l@{}}PreAct-\\ResNet18\end{tabular} & MIT  & 50K/10K   & \begin{tabular}[c]{@{}l@{}}1 Target Class Spread  \\ Into 9 Other Classes\end{tabular}                      \\

\midrule
\begin{tabular}[c]{@{}l@{}}Runtime \\VS Perf (Fig~\ref{fig:runtime})\end{tabular}    & CIFAR10   & MIT       & ResNet18   & MIT  & 2K/1K   & 10\% Backdoors    \\

\midrule
\begin{tabular}[c]{@{}l@{}}Backdoor: \\Blend (Fig~\ref{fig:back-gtsrb})\end{tabular}   & GTSRB  & CC0 1.0  & \begin{tabular}[c]{@{}l@{}}PreAct-\\ResNet18\end{tabular} & MIT  & \begin{tabular}[c]{@{}l@{}}35888/\\12360\end{tabular} & 5\% Poisoned \\

\midrule
Noisy Feature (Fig~\ref{fig:back-gtsrb})& MNIST  & \begin{tabular}[c]{@{}l@{}}CCA-\\SA 3.0\end{tabular} & \begin{tabular}[c]{@{}l@{}}PreAct-\\ResNet18\end{tabular} & MIT & 60K/10K & 25\% Noisy Data                                                                         \\

\midrule
Time Complexity  (Table~\ref{table:time-scale})   & \begin{tabular}[c]{@{}l@{}}ImageNet\\-100\end{tabular} & \cellcolor[HTML]{C0C0C0}{\color[HTML]{FFFFC7} Not Applicable} & ResNet50   & MIT  & 100K/10K  & \begin{tabular}[c]{@{}l@{}}25\% Mislabeled\\ 10 Classes\end{tabular} \\

\midrule
Irrelevant Data  (Fig~\ref{fig:irrel})   & CIFAR100  & MIT  & \begin{tabular}[c]{@{}l@{}}PreAct-\\ResNet34\end{tabular} & MIT    & 50K/10K  & \begin{tabular}[c]{@{}l@{}}1 Target Class Spread \\Into 99 Other Classes\end{tabular}  \\ 

\midrule
Noisy Label   (Fig~\ref{fig:mislabel-fe}) & CIFAR10    & MIT    & \begin{tabular}[c]{@{}l@{}}PreAct-\\ResNet18\end{tabular} & MIT & 50K/10K   & 25\% Noisy Label                                                                        \\
  & &  & VGG16  & \begin{tabular}[c]{@{}l@{}}CC \\BY 4.0\end{tabular}   &   &  \\
  & &  & ResNet18   & MIT   &   &     \\
  
  \midrule
Noisy Label   (Fig~\ref{fig:misl-lbl-wgt})& CIFAR10    & MIT   & \begin{tabular}[c]{@{}l@{}}PreAct- \\ResNet18\end{tabular} & MIT  & 50K/10K  & \begin{tabular}[c]{@{}l@{}}Different Feature \\Weights 1,5,10,100\end{tabular}  \\
  &     &       &   &     &   & \begin{tabular}[c]{@{}l@{}}Different Label\\Weights 1,10,50,100\end{tabular}            \\
  &    &     &     &     & \begin{tabular}[c]{@{}l@{}}50K/10K\\50K/5K\\50K/2K\\50K/0.5K\\50K/0.2K\end{tabular} & \begin{tabular}[c]{@{}l@{}}Different Validation\\Sizes\end{tabular}                     \\
  
  \midrule
\begin{tabular}[c]{@{}l@{}}Backdoor: Blend,\\Trojan, SQ-WM (Fig~\ref{fig:back_grad})\end{tabular}      &    CIFAR10  &   MIT    &  \cellcolor[HTML]{C0C0C0}{\color[HTML]{FFFFC7} Not Applicable}  &    \cellcolor[HTML]{C0C0C0}{\color[HTML]{FFFFC7} Not Applicable} & 50K/10K   & \begin{tabular}[c]{@{}l@{}}Visualization of Attacks\end{tabular}      
\\
   &      &         &       &     &     & \begin{tabular}[c]{@{}l@{}}Detection Rate of $\AlgName$\end{tabular}  \\
   
   \midrule
Data Duplication     (Table~\ref{table:dup}) &          \cellcolor[HTML]{C0C0C0}{\color[HTML]{FFFFC7} Not Applicable}       &      \cellcolor[HTML]{C0C0C0}{\color[HTML]{FFFFC7} Not Applicable}  &     \cellcolor[HTML]{C0C0C0}{\color[HTML]{FFFFC7} Not Applicable}   &     \cellcolor[HTML]{C0C0C0}{\color[HTML]{FFFFC7} Not Applicable}    & 5K/5K     & \begin{tabular}[c]{@{}l@{}}Duplication of\\Training Set   \end{tabular}
\\

\midrule
Dataset Reduction  (Fig~\ref{fig:reduce}) & CIFAR10 & MIT  & \begin{tabular}[c]{@{}l@{}}PreAct-\\ResNet18\end{tabular} & MIT  & 25K/10K   & 2.5K Samples From Each Class \\

\midrule
Data Summarization (Fig~\ref{fig:addition})& CIFAR10 & MIT  & \begin{tabular}[c]{@{}l@{}}PreAct-\\ResNet18\end{tabular} & MIT  & 25K/10K   & 2.5K Samples From Each Class \\

\midrule
Unbalanced Dataset (Fig~\ref{fig:unbalance})& CIFAR10 & MIT  & \begin{tabular}[c]{@{}l@{}}PreAct-\\ResNet18\end{tabular} & MIT  & 27.5K/10K   & \begin{tabular}[c]{@{}l@{}}5K Samples From Class Frog  \\ 2.5K Samples From Other Classes\end{tabular} \\

\midrule
Time Complexity (Fig~\ref{fig:nearlinear})& CIFAR10 & MIT  & \begin{tabular}[c]{@{}l@{}}PreAct-\\ResNet18\end{tabular} & MIT  & 50K/10K   & 5K From Each Class \\
\bottomrule
\end{tabular}
\end{adjustbox}
\end{table}

% \cellcolor[HTML]{C0C0C0}{\color[HTML]{FFFFC7} Not Applicable}

\textbf{Hardware.} A server with an NVIDIA Tesla P100-PCIE-16GB graphic card is used as the hardware platform in this work. 
% Particularly, each experiment takes within one GPU hour of compute. 

\textbf{Software.} 

\begin{wraptable}{r}{0.35\columnwidth}
 \vspace{-2em}
\begin{tabular}{r|rr}
\multicolumn{1}{c}{\textbf{Dataset}}                      & \multicolumn{2}{c}{\textbf{ OT Dist}}  \\
\multicolumn{1}{c}{\textbf{Size}} & Direct  & Near                      \\ 
\toprule
$5000          $                             & $195.64$ & $+0.98$                                     \\
$2 \times 5000    $                               & $+0.00$ & $+0.98$                                     \\
$3 \times 5000 $                                  & $+0.00$ & $+0.98$                                     \\
$4 \times 5000  $                                 & $+0.00$ & $+0.98$                                     \\
$4.5 \times 5000  $                               & $+0.07$ & $+0.98$                                     \\
$5 \times 5000    $                               & $+0.00$ & $+0.98$                                    
\end{tabular}
% \vspace{1em}
\caption{Class-wise Wasserstein distance behavior under dataset direct duplication and its near duplicates.}%\yi{re-scale your table to mach the text size of Figure 2}}
\label{table:dup}

\end{wraptable}

For our implementation, we use \texttt{PyTorch} for the main framework~\citep{NEURIPS2019_9015}, assisted by three main libraries, which are \texttt{otdd} (optimal transport calculation setup with datasets)~\citep{alvarez2020geometric}, \texttt{geomloss} (actual optimal transport calculation)~\citep{feydy2019interpolating}, and \texttt{numpy} (tool for array routines)~\citep{harris2020array}.

\end{appendices}

\end{document}